\documentclass[manuscript,screen]{acmart}

\usepackage{multirow}
\usepackage{ragged2e}
\usepackage{lineno,hyperref}
\usepackage{setspace}
\usepackage{booktabs}
\usepackage{subfigure}
\usepackage{graphicx}
\usepackage{color,xcolor}
\def\degree{${}^{\circ}$}

\usepackage{setspace}

\AtBeginDocument{%
  \providecommand\BibTeX{{%
    \normalfont B\kern-0.5em{\scshape i\kern-0.25em b}\kern-0.8em\TeX}}}

\setcopyright{acmcopyright}
\acmJournal{CSUR}
\acmYear{2022} \acmVolume{1} \acmNumber{1} \acmArticle{1} \acmMonth{1} \acmPrice{15.00}\acmDOI{10.1145/3524496}

\usepackage{color}

\begin{document}

\title{Deep Learning on Monocular Object Pose Detection and Tracking: 
A Comprehensive Overview}

\author{Zhaoxin Fan}
\email{fanzhaoxin@ruc.edu.cn}
\affiliation{%
  \institution{Key Laboratory of Data Engineering and Knowledge Engineering of MOE, School of Information, Renmin University of China, Beijing, China}
  \streetaddress{No. 59, Zhongguancun Street, Haidian Dist.}
  \city{Beijing}
  \country{P.R.China}
  \postcode{100872}}

\author{Yazhi Zhu}
\email{19120324@bjtu.edu.cn}
\affiliation{%
  \institution{Institute of Information Science, Beijing Jiaotong University}
  \streetaddress{No.3 Shangyuancun, Haidian Dist.}
  \city{Beijing}
  \country{P.R.China}}

\author{Yulin He}
\email{yolanehe@ruc.edu.cn}
\affiliation{%
	\institution{Key Laboratory of Data Engineering and Knowledge Engineering of MOE, School of Information, Renmin University of China, Beijing, China}
	\streetaddress{No. 59, Zhongguancun Street, Haidian Dist.}
	\city{Beijing}
	\country{P.R.China}
	\postcode{100872}}

\author{Qi Sun}
\email{ssq@ruc.edu.cn}
\affiliation{%
	\institution{Key Laboratory of Data Engineering and Knowledge Engineering of MOE, School of Information, Renmin University of China, Beijing, China}
	\streetaddress{No. 59, Zhongguancun Street, Haidian Dist.}
	\city{Beijing}
	\country{P.R.China}
	\postcode{100872}}

\author{Hongyan Liu}
\email{liuhy@sem.tsinghua.edu.cn}
\affiliation{%
  \institution{School of Economics and Management,  Tsinghua University}
  \streetaddress{Haidian Dist}
  \city{Beijing}
  \country{P.R.China}
  \postcode{100084}}

\author{Jun He}
\email{hejun@ruc.edu.cn}
\affiliation{%
	\institution{Key Laboratory of Data Engineering and Knowledge Engineering of MOE, School of Information, Renmin University of China, Beijing, China}
	\streetaddress{No. 59, Zhongguancun Street, Haidian Dist.}
	\city{Beijing}
	\country{P.R.China}
	\postcode{100872}}

\thanks{Corresponding author: J. He.}

\renewcommand{\shortauthors}{Zhaoxin Fan, et al.}

\begin{abstract}
Object pose detection and tracking has recently attracted increasing attention due to its wide applications in many areas, such as autonomous driving, robotics, and augmented reality. Among methods for object pose detection and tracking, deep learning is the most promising one that has shown better performance than others. However, survey study about the latest development of deep learning-based methods is lacking. Therefore, this study presents a comprehensive review of recent progress in object pose detection and tracking that belongs to the deep learning technical route.  To achieve a more thorough introduction, the scope of this study is limited to methods taking monocular RGB/RGBD data as input and covering three kinds of major tasks: instance-level monocular object pose detection, category-level monocular object pose detection, and monocular object pose tracking.  In our work, metrics, datasets, and methods of both detection and tracking are presented in detail.  Comparative results of current state-of-the-art methods on several publicly available datasets are also presented, together with insightful observations and inspiring future research directions. 
\end{abstract}


\maketitle

\section{Introduction}
Object pose detection and tracking is critical problems that has been extensively studied in the past decades. It is closely related to many rapidly evolving technological areas such as autonomous driving, robotics, and augmented reality (AR).

For autonomous driving \cite{maurer2016autonomous,levinson2011towards,wang2018networking,grigorescu2020survey},  the task is often called 3D object detection (tracking).  It can provide self-driving cars accurate positional and orientational information of other objects on the road, thus helping cars avoid possible collisions or accidents in advance.  For robotics \cite{tremblay2018deep,bousmalis2018using,james2019sim,morrison2018closing}, the estimated object pose can provide the robot positional information for grasping, navigation, and other actions related to environmental interaction.  For AR \cite{ibanez2018augmented,cipresso2018past,gattullo2019towards,peddie2017augmented}, the estimated object pose can provide sufficient information for 2D 3D interaction.  For example, projection from the real world to the image can be achieved using the estimated object pose; thus, further actions such as special effect injection can be done.

Table \ref{Tasks_we_care} shows tasks concentrated on in this survey. Specifically, this survey covers two major tasks: monocular object pose detection and monocular object pose tracking. Both are introduced from the perspective of instance-level and category-level. Hence, these two major tasks can be divided into subtasks: instance-level object pose detection, category-level object pose detection, instance-level object pose tracking, and category-level object pose tracking. However, related works are few for tracking tasks, as the field is in a relatively immature development stage. Moreover, the outputs of the two tasks are unified.  Therefore, the two tasks are combined as a unified task in our survey. Unless otherwise specified, \emph{objects} in this survey refers to rigid objects.   

These four topics are not independent. Specifically, first, all of them are targeted at predicting the 6D pose of a rigid object. Second, the category-level object pose detection/tracking is a more general extension of instance-level object pose detection/tracking. Therefore, they are high-related tasks in expected output and technical routes. Third, pose detection and pose tracking are complementary.Pose detection results could serve as initialization of tracking methods, and pose tracking is often used to improve the robustness and the smoothness of detection methods. This paper provides a comprehensive overview of several highly related topics.

\begin{table*}[]
	\centering
		\begin{tabular}{ r|c|c|c }
			\hline
			{\bf Tasks} & {\bf Subtasks} & {\bf Input} & {\bf Output} \\
			\hline
			\multirow{ 2}{3cm}{Monocular object pose detection} & Instance-level object pose detection & RGB(D) image+CAD   & $\mathcal{R}$ , $\mathcal{T}$ \\
			\cline{2-4}
			\multicolumn{ 1}{ c|}{} & Category-level object pose detection & RGB(D) image & $\mathcal{R}$ , $\mathcal{T}$ , $\mathcal{S}$ \\
			\hline
			\multirow{2}{3cm}{Monocular object pose tracking} & Instance-level object pose tracking & RGB(D) video+initial pose+CAD  & $\mathcal{R}$ , $\mathcal{T}$ \\
			\cline{2-4}
			\multicolumn{ 1}{c|}{} & Category-level object pose tracking & RGB(D) video+initial pose & $\mathcal{R}$ , $\mathcal{T}$ \\
			\hline
	\end{tabular} 
	\caption{Tasks we consider in this survey. Here, $\mathcal{R}$, $\mathcal{T}$ and $\mathcal{S}$ denote rotation, tranlslation and object size, respectively.}
	\label{Tasks_we_care}
	\vspace{-0.35in}
\end{table*}

In the progress of our research, several existing survey works about Object Pose Detection are found\cite{sahin2018recovering,patil2018survey,du2019vision,kleeberger2020survey,chen2020survey,sahin2020review}. The differences between these works and this paper include the following aspects. 1) Previous works involve traditional and deep learning methods, whereas this work mainly focuses on researching and analyzing the deep learning models in a very detailed manner. 2) Previous works review methods take all kinds of input data formats as input, whereas this work only introduces monocular methods in a fine-grained way. 3) Several of the previous works do not introduce task definition and metrics definition, which would make the scope of their surveys obscure. Our work defines them in detail to provide readers with a clear understanding of this field. 4) Few previous works introduce current large-scale datasets in detail. However, datasets play a crucial role in deep learning. Therefore, a comprehensive description and comparison of existing open-source datasets are presented in this survey. 5) Previous works do not review works concerning object pose tracking. This work argues that pose detection and pose tracking are highly related. Hence, analyzing them in a unified work is necessary. 6) Few previous works analyze current challenges and possible future works. This paper provides suggestions for future works and corresponding reasons after introducing existing works.

\begin{figure*}
	\centering  
	\includegraphics[width=\textwidth]{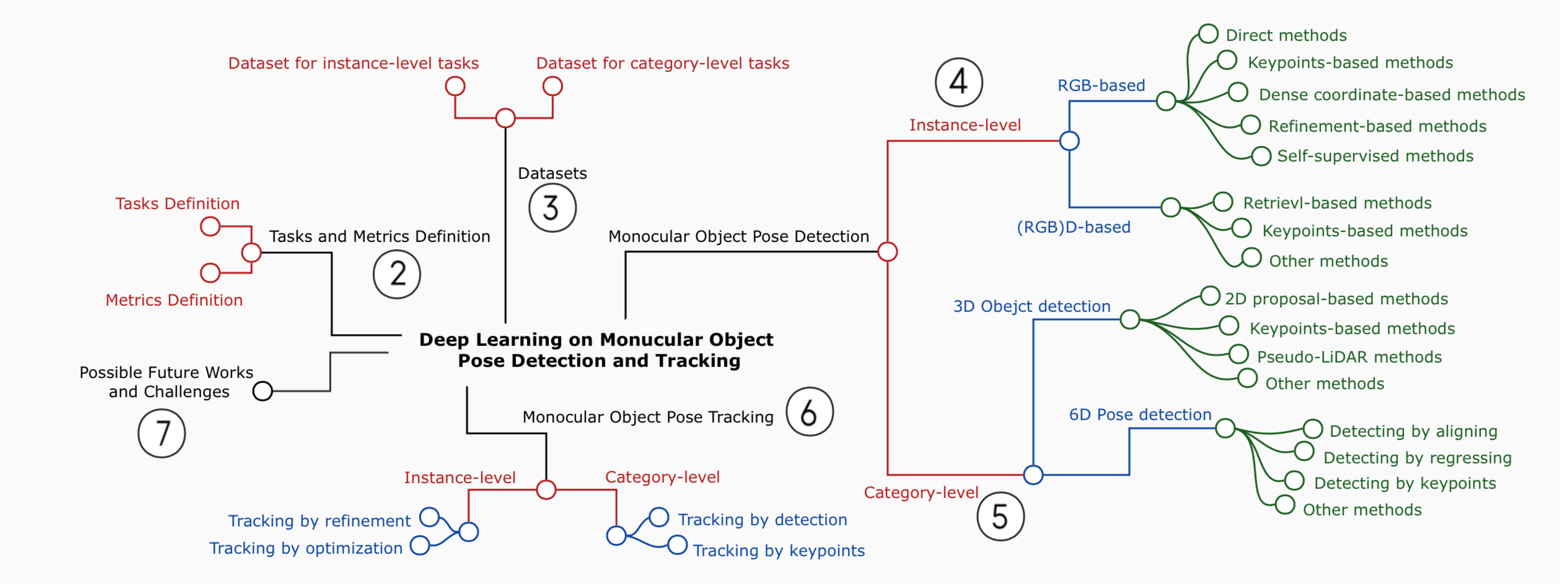}  
	\vspace{-0.3in} 
	\caption{A taxonomy of deep learning methods for monocular object pose detection and tracking}
	\label{overview}
	\vspace{-0.2in}
\end{figure*}

\textbf{Contribution:} The major contributions of this work are summarized as follows:
\begin{itemize}
	\item To the best of our knowledge, this is the first survey that comprehensively covers task definition, metrics definition, datasets, and intact detailed illustration of deep learning-based object pose detection and tracking methods with monocular RGB/RGBD input. In contrast to existing reviews, this work focuses on deep learning methods for monocular RGB/RGBD image input rather than all types of data format. 
	
	\item This study addresses the most recent and advanced progress of object pose detection and tracking. Thus, it provides readers an insight into the most cutting-edge technologies, which benefits the academia and the industry. 
	
	\item This work presents a full comparison, qualitative and quantitative, and analysis of state-of-the-art methods. Hence, the advantages and the disadvantages of different techniques are clear at a glance. 
	
	\item This work provides a comprehensive analysis for potential future research directions and challenges, which may be helpful for readers to find possible breakthrough points that may benefit their future works from different perspectives. 
\end{itemize}

\textbf{Organization:} Figure \ref{overview} illustrates a system flowchart of the topics focused in this survey to provide readers an overview of our work. This figure also indicates the organization of this survey. Specifically, as mentioned above, the monocular object pose detection and tracking task can be classified into four subtasks, among which instance-level object pose tracking and category-level object pose tracking are combined as a unified task for better literature classification and better logical organization.  Instance-level monocular object pose detection, category-level monocular object pose detection, and monocular object pose tracking are defined as three horizontal tasks in this survey. Therefore, three kinds of tasks are organized and introduced in our work, as shown in Figure \ref{overview}. 

First, we introduce the detailed definitions of the three tasks and several most commonly used metrics for evaluating object pose detection (tracking) algorithms are introduced in Section \ref{definiton}. Datasets used to train and test deep learning models are presented and compared in Section \ref{secdataset}. Techniques about the three tasks are reviewed in detail in  Sections \ref{secinstance}, \ref{seccatefory} and \ref{sectracking} with fine-grained classifications. A complete comparison of these methods is also presented qualitatively and quantitatively. Finally, future research directions and challenges are analyzed in Section \ref{secfuturework}, and our work is concluded in Section \ref{secconclusion}. 

\begin{figure}[ht]
	\centering  
	\includegraphics[width=1.0\textwidth]{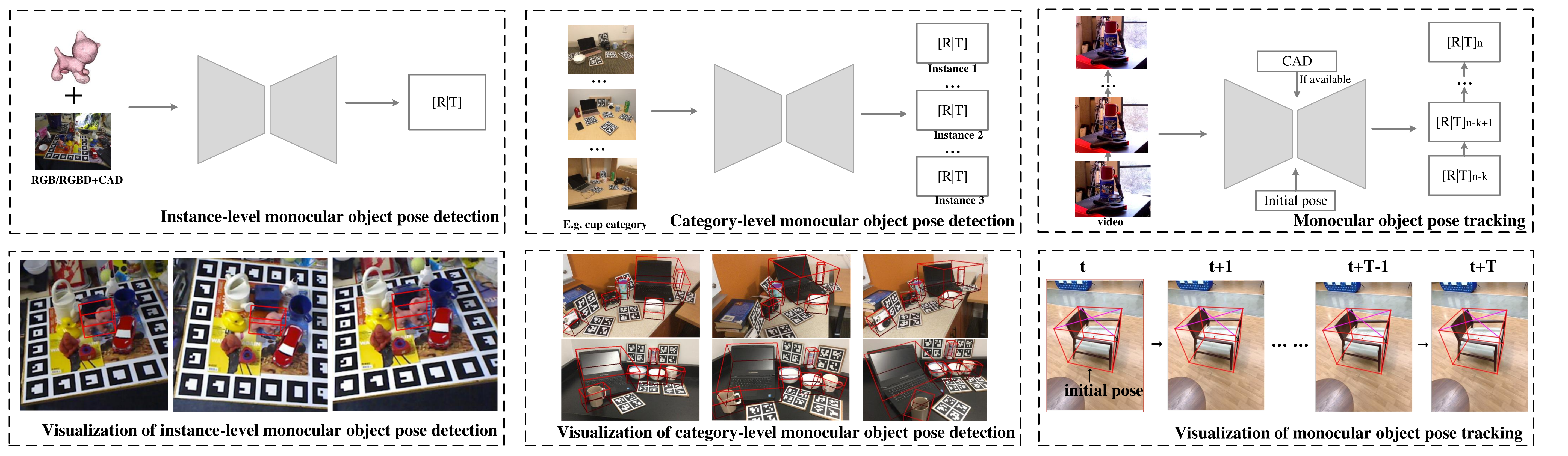}  
	\vspace{-0.3in} 
	\caption{Illustrations of tasks involved in this survey. The bottom row show some visualization results of expected output. For instance-level object pose detection, the 6Dof pose of a particular object(a cat in the figure) is expected. For category-level object pose detection, the  9Dof poses of a category of objects are expected. For object pose tracking, given the initial pose, the 6Dof poses of the target object in the following frames are expected.} 
	\label{task}
	\vspace{-0.2in} 
\end{figure}

\section{Tasks and Metrics Definition}
\label{definiton}
Before reviewing relevant works, the three tasks discussed in this study are first defined, and several commonly used metrics are introduced in this section. Figure \ref{task} top shows illustrations of the three tasks. Figure \ref{task} bottom shows several visualization examples of expected output of each task. Table \ref{Tasks_we_care} shows an overview of their input and output.

\subsection{Task Definition}
\textbf{Instance-level monocular object pose detection}: Given a monocular RGB/RGBD image $\mathcal{I}$ and a CAD model $\mathcal{M}$  of a target instance, the object pose $\mathcal{P} \in SE(3)$ of the target instance observed needs to be recovered in the image, where  $\mathcal{P}$ can be decoupled to rotation $\mathcal{R} \in SO(3)$ and translation $\mathcal{T} \in R^3$ w.r.t the camera. Rotation $\mathcal{R}$ consists of pitch, yaw and roll, which refer to rotations around the X, Y, and Z axes respectively. They constitute the object's rotation around the overall 3D space. Translation $\mathcal{T}$ refers to the x, y, and z components of the object center in the camera coordinate system.  $\mathcal{P}$ can also be understood from another perspective as a function of transforming an object from its object coordinate system to the camera coordinate system. The whole task can be expressed as:
\begin{quotation}
	\centering
	$\mathcal{[R|T]}$ =$\mathcal{F\{[I,M]| \theta\}}$
\end{quotation}
where  $\mathcal{F}$ refers to a specific deep learning model, and $\mathcal{\theta}$ refers to the model's parameters.  In the task, rotation and translation together form 6 degrees of freedom (6Dof). A well-trained model only works for detecting the pose of a specific object instance.

\textbf{Category-level monocular object pose detection}: Given a monocular RGB/RGBD image $\mathcal{I}$, the pose $\mathcal{P} \in SE(3)$ of the target object in the image needs to be recovered with no CAD model available. Object size $\mathcal{S} \in R^3$ (i.e. object length, width and height) also needs to be estimated. In the category-level monocular object pose detection task, a well-trained model should obtain the ability to recognize a line of object instances that belong to the same category.  Rotation, translation and object size together form 9 degrees of freedom (9Dof). Thus, strictly speaking, the task should be defined as category-level monocular 9Dof pose detection.  In addition, in several works, especially in an autonomous driving scenario, only the 7Dof pose is required to be predicted, where pitch and roll are excluded. The task can be expressed as:
\begin{quotation}
	\centering
	$\mathcal{[R|T|S]}$ =$\mathcal{F\{[I]| \theta\}}$
	
\end{quotation}
where  $\mathcal{F}$ refers to a specific deep learning model and $\mathcal{\theta}$ refers to the model parameters.

\textbf{Monocular object pose tracking}: Given a series of monocular RGB/RGBD images $\mathcal{I}_{n-k}$, $\mathcal{I}_{n-k+1}$,..., $\mathcal{I}_n$ from time steps $T_{n-k}$ to $T_{n}$ and initial pose $\mathcal{P}_0$ of the target object, poses of the target object at all time steps need to be recovered. The task at time step $T_n$ can be expressed as:
\begin{quotation}
	\centering
	$\mathcal{P}_n$ =$\mathcal{F}\{\mathcal{I}_{n},...,\mathcal{I}_{n-k+1},\mathcal{I}_{n-k}; \mathcal{P}_0| \theta \}$
	
\end{quotation}
where  $\mathcal{F}$ refers to the tracking algorithm and $\mathcal{\theta}$ refers to the model parameters. The task can also be divided into instance-level object pose tracking and category-level object pose tracking. In both situations, $\mathcal{P}_n$ means rotation $\mathcal{R} \in SO(3)$ and translation $\mathcal{T} \in R^3$. Object size does not need to be estimated because it can be directly inferred from the initial pose $\mathcal{P}_0$. Similarly, in several autonomous driving-related works, pitch and roll do not need to be estimated. During tracking, the CAD models of the target instances can be used in the instance-level object  pose tracking task but cannot be used in the category-level object pose tracking task.

\subsection{Metrics Definition}
\textbf{2D Projection metric} \cite{brachmann2016uncertainty} computes the mean distance between the projections of CAD model points after being transformed by the estimated object pose and the ground-truth pose. A pose is considered correct if the distance between two projections is less than 5 pixels. This metric is ideal for evaluating AR-related algorithms. However, the 2D projection metric is only suitable for evaluating instance-level pose detection or tracking methods because the CAD model of the target object is required to compute projections.

\textbf{ n\degree n cm metric } \cite{shotton2013scene} measures whether the rotation error is less than
n\degree and whether translation error is below n cm. If both are satisfied, the pose is considered correct. For symmetrical objects, n\degree n cm is computed as the slightest error for all possible ground-truth poses. The most commonly used threshold settings include 5\degree 5 cm, 5\degree 10 cm and 10\degree 10 cm. This metric is suitable for evaluating instance-level methods and category-level methods. Mean average precision (mAP) at n\degree n cm (mAP@n\degree n cm) is often reported in category-level methods.

\textbf{ADD metric} \cite{hodavn2016evaluation,hinterstoisser2012model} computes the mean distance between two transformed model points using the estimated pose and the ground-truth pose. When the distance is less than 10\% of the model’s diameter, the estimated pose is correct. For symmetric objects, the ADD metric is often replaced by the ADD-S metric, where the mean distance is computed-based on the closest point distance. When models trained on datasets such as YCB video \cite{xiang2017posecnn} are evaluated, the ADD(-S) AUC (Area Under Curve) should also be computed as a metric. ADD(-S) is only suitable for evaluating instance-level object pose detection or tracking methods.

\textbf{nuScenes detection score (NDS) metric} \cite{caesar2020nuscenes} calculates a weighted sum  of the \emph{mAP and the set of the five mean True Positive metrics.} This metric is proposed to reflect the  velocity and attribute estimation accuracy in the nuScenes dataset. Thus, half of NDS is thus based on the detection performance, whereas the other half quantifies the quality of the detections in terms of box
location, size, orientation, attributes, and velocity. 

\textbf{3D Intersection over Union (3DIoU) metric} \cite{geiger2012we, wang2019normalized,ahmadyan2020objectron} , also known as 3D Jaccard index, measures how close the predicted bounding boxes are to the ground truths. Here, the predicted bounding boxes and the ground truths can be easily computed using the predicted and ground truth object poses along with the object size or the CAD model. 3DIoU takes the predicted box and the ground truth box as input and computes the intersection volume of the two boxes. The final 3DIoU value is equal to the volume of the intersection of two boxes divided by the volume of their union. Therefore,  3DIoU is a normalized metric, and its value ranges from 0 to 1.  3DIoU is suitable for evaluating instance-level methods and category-level methods. In most of the existing works,  mAP@IoU25, mAP@IoU50 and  mAP@IoU75, that is, if the 3DIoU is higher than 0.25, 0.5 or 0.75, respectively, the predicted pose is regarded as correct, are the most frequently reported indicators.

\section{Datasets for Monocular Object Pose Detection and  Tracking}
\label{secdataset}

\begin{table*}
	\centering
	\resizebox{\textwidth}{15mm}{
		\begin{tabular}{   r | c | c | c | c | c | c | c | c  }
			\hline
			Datasets 	& Levels &Format	& Num\_Categories & CAD models&Annotation &Resolution&Scale&Usage\\ 
			\hline
			\textbf{Linemod}\cite{hinterstoisser2012model} 	& Instance &RGBD	& 13 & Yes&6Dof &$640 \times 480$& 15k&detection\\ 
			\hline 
			\textbf{Linemod Occlusion}\cite{brachmann2014learning} & Instance &RGBD	& 8 & Yes&6Dof &$640 \times 480$& 1.2k&detection \\ 
			\hline 
			\textbf{YCB video} \cite{xiang2017posecnn} & Instance &RGBD	& 21 & Yes&6Dof &$640 \times 480$& 134k &detection \& tracking\\ 
			
			\hline 
			\textbf{T-LESS} \cite{hodan2017t}  & Instance &RGBD	& 30 & Yes&6Dof &-& 49k &detection \\ 
			\hline 
			
			\textbf{HomebrewedDB} \cite{kaskman2019homebreweddb}  & Instance &RGB	& 33 & No&6Dof &-&17k  &detection \& tracking\\ 
			\hline 
			\textbf{KITTI3D} \cite{geiger2012we} & Category &RGB	& 8 & No&7Dof &$384 \times 1280$&15k  &detection\\ 
			\hline
			
			\textbf{Apolloscape} \cite{huang2018apolloscape} & Category &RGB	& 3 & Yes&9Dof &$2710 \times 3384$&5k  &detection\\ 
			
			\hline
			
			\textbf{Nuscenes} \cite{caesar2020nuscenes} & Category &RGBD	& 23 & No&7Dof &$1600 \times 900$& 40k  &detection \& tracking\\ 
			\hline
			\textbf{NOCS(CAMERA275)}\cite{wang2019normalized} & Category &RGBD	& 6 & Yes&9Dof &$640 \times 480$&300k  &detection\\ 
			
			\hline
			
			\textbf{NOCS(REAL25)}\cite{wang2019normalized} & Category &RGBD	& 6 & Yes&9Dof &-&8k  &detection\\ 
			\hline
			
			\textbf{Objectron} \cite{ahmadyan2020objectron}  & Category &RGB	& 9 & No&9Dof &$1020 \times 1440$&4M  &detection \& tracking\\ 
			\hline 
	\end{tabular}}
	\caption{A comparison of datasets for monocular object pose detection and tracking.}
	\label{dataset}
	\vspace{-0.35in} 
\end{table*}

\begin{figure*}[t]
	\centering  
	\includegraphics[width=0.95\textwidth]{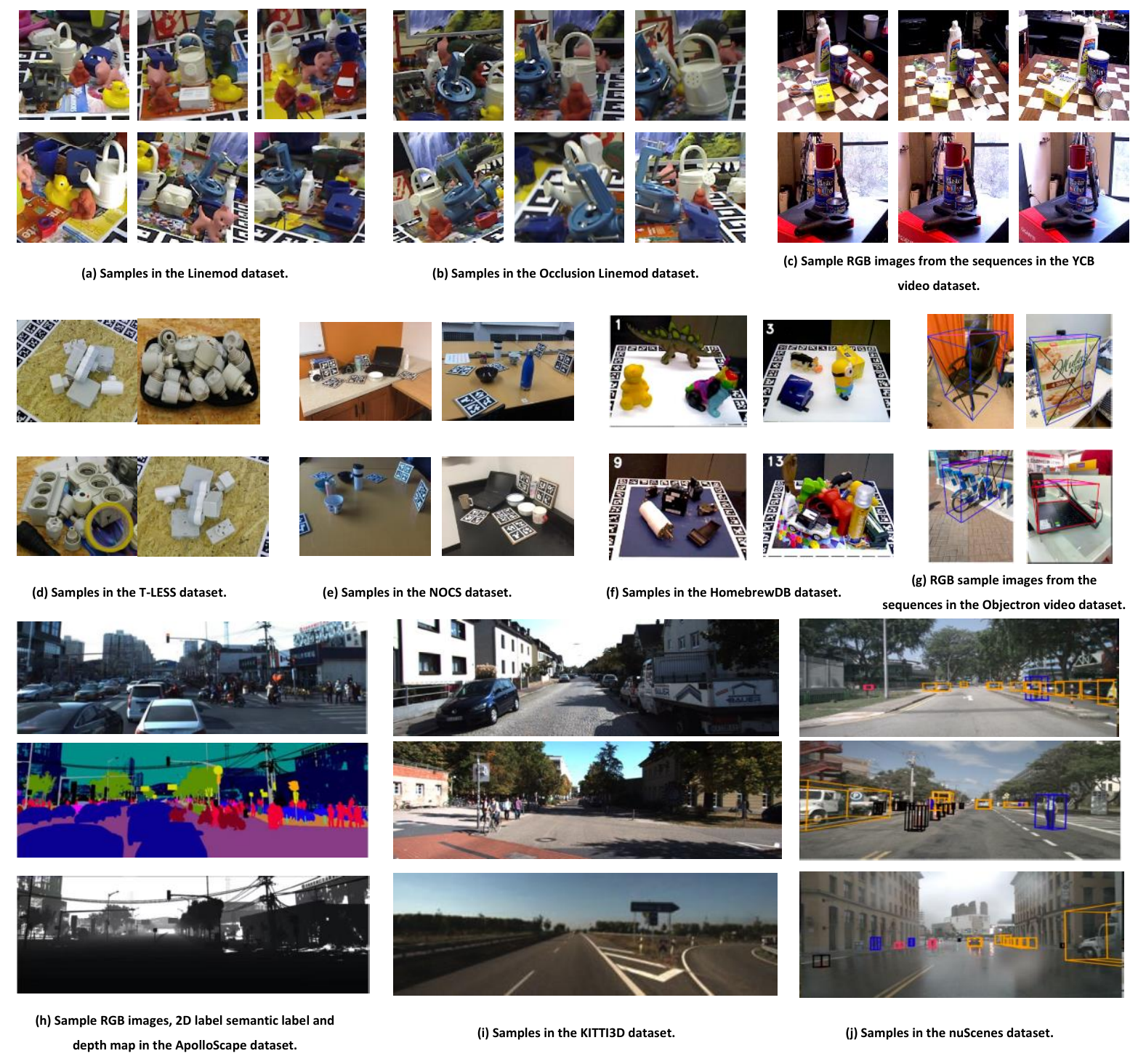}
	\vspace{-0.2in}   
	\caption{Image examples of commonly used datasets.} 
	\label{posedatasets}
	\vspace{-0.28in} 
\end{figure*}

To date, object pose detection and tracking have developed into a large, important research field. The progress of deep learning-based object pose detection and tracking is inseparable from the use and construction of many large-scale datasets that are both challenging and credible. In this section, commonly used mainstream monocular object pose detection and tracking-related datasets are introduced. Here, they can be categorized into ones for instance-level object pose detection/tracking and ones for category-level object pose detection/tracking according to whether CAD models of objects are available and whether there exists multiple instances exist for each category.  Table \ref{dataset} shows an overview and a full comparison of these datasets, and Figure \ref{posedatasets} exhibits visualization examples.

\subsection{Datasets for Instance-Level Object Pose Detection and Tracking}
\textbf{Linemod} \cite{hinterstoisser2012model} is the most standard benchmark for instance-level object pose detection and consists of 13 RGBD sequences. Each sequence contains around 1.2k images with ground-truth 6Dof poses (3 degrees for rotation and 3 degrees for translation) of each target object. The resolution of each image is $640 \times 480$. For most of the studies, around 15\% of images of each sequence are selected for training, and the remaining 85\% are selected for testing. Each sequence includes challenging cluttered scenes, textureless objects, and light condition variations, bringing difficulties for accurate object pose prediction. Though containing sequences, the dataset is always only used for object pose detection rather than tracking, as the sequences are short and the train/test split has made them noncontinuous.

\textbf{Linemod Occlusion} \cite{brachmann2014learning} is generated from Linemod dataset to compensate for its lack of occlusion cases. It consists of 1214 RGBD images belonging to \emph{benchvise} sequence of Linemod, where ground-truth 6Dof poses of another eight severely occluded visible objects are annotated additionally. In most works, Linemod Occlusion is only used for testing, and the models are trained on Linemod. 

\textbf{YCB video} \cite{xiang2017posecnn} consists of 21 objects collected from YCB object set \cite{calli2015ycb}. The 21 objects are selected due to their high-quality 3D models and good visibility in depth. A total of 92 videos are in the entire YCB video dataset, and each video is composed of three to nine YCB objects, which are captured by a real RGBD camera. All videos in the dataset consist of 133,827 images. The resolution of each image is $640 \times 480$.  Varying lighting conditions, substantial image noise, and severe occlusions make detecting object pose in this dataset challenging.  Given that it is made up of videos, this dataset can be used for instance-level object pose detection and instance-level object pose tracking. Commonly, 80 videos are used for training, and  2,949 keyframes extracted from the rest 12 videos are used for testing.

\textbf{T-LESS} \cite{hodan2017t} is an RGBD dataset with textureless objects. The dataset consists of 30 electrical parts with no evident texture, distinguishable color, or unique reflection characteristics, and usually have similar shapes or sizes. Images in T-LESS are captured by three synchronized sensors: a structured light sensor, an RGBD sensor, and a high-resolution RGB sensor. It provides 39K RGBD images for  training  and 10K for testing. Images with different resolutions are provided in the dataset. Specifically, there are three kinds of resolution ($400 \times 400, 1900 \times 1900, and 720 \times 540$) are for RGBD images and one kind of resolution ($2560 \times 1920$) is for RGB images. Backgrounds of images in the training set are mostly black, whereas the backgrounds in the test set vary much, including different lighting and occlusion. The texturelessness of objects and the complexity of environments make it a challenging dataset.

\textbf{HomebrewedDB} \cite{kaskman2019homebreweddb} contains 33 highly accurately reconstructed 3D models of toys, household objects and low-textured industrial objects  (specifically, 17 toys, 8 household objects, and 8 industry-relevant objects) of sizes varying from 10.1 cm to 47.7 cm in diameter. The dataset consists of 13 sequences from different scenes, each containing 1340 frames filmed using two different RGBD sensors. All data are collected from scenes that span a range of complexity from simple (3 objects on a plain background) to complex (highly occluded with 8 objects and extensive clutter). Precise 6Dof pose annotations are provided for dataset objects in the scenes, obtained using an automated pipeline. Resolutions of images are quite different in several ways because data are collected by different sensors. For instance, color and depth images in the sequences recorded by Carmine sensor have a default resolution of $640 \times 480$, whereas images collected by Kinect maintain a size of $1920 \times 1080$ in the final.  Texturelessness, scalability changes, occlusion changes, light condition changes, and object appearance changes together constitute this dataset's complexity.

\subsection{Dataset for Category-Level Object Pose Detection and Tracking}
\textbf{KITTI3D} \cite{geiger2012we} is the most widely used dataset for category-level monocular object pose detection. The most commonly used data split is 7481 RGB images for training and 7518  RGB images for testing. All images are annotated with 7 degree of freedom (7Dof) labels: three for object translation, three for object size, and one for object rotation. A total of eight object categories are annotated, and the three most commonly evaluated categories are \emph{car, pedestrian,  and cyclist}. The resolution of each image is $384 \times 1280$. Strictly speaking, due to the lack of annotations for pitch and roll, the 9Dof pose of objects cannot be fully recovered using models trained on this dataset. However, this dataset has received extensive attention because its application scenario is about autonomous driving, and 7Dof information is rich enough for this task. In KITTI3D, only annotations of training split are publicly available, and evaluation scores on test split can only be obtained by submitting prediction results to the official benchmark website. Each image in KITTI3D is provided a corresponding point cloud, collected by a LiDAR, which is broadly used for 3D object detection. Works about utilizing point clouds for 3D object detection on KITTI3D will not be reviewed because this study only reviews works for monocular object pose detection.

\textbf{Apolloscape} \cite{huang2018apolloscape} is originally released for the \emph{ Apolloscape 3D Car Instance challenge}. The released dataset contains a diverse set of stereo video sequences recorded in street scenes from different cities. A total of 3,941, 208, and 1,041 high-quality annotated images are in the training, validation, and test sets, respectively. The resolution of each monocular RGB image is $2710 \times 3384$. The annotations are 9 degrees of freedom (9Dof). Therefore, compared with KITTI3D, Apolloscape is more suitable for evaluating performances of category-level object pose detection and tracking-related models. Another advantage of this dataset is that the car models are provided. In total, 79 car models belonging to three categories (sedan1, sedan2, and SUV) are provided in the whole dataset, and only 34 models appear in the training set.

\textbf{Nuscenes} \cite{caesar2020nuscenes} is a large-scale public dataset for autonomous driving. It contains 1000 driving scenes in Boston and Singapore. The rich complexity of nuScenes encourages the development of methods that enable safe driving in urban areas with dozens of objects per scene. The full dataset includes approximately 1.4M camera images, 390k LIDAR sweeps, 1.4M RADAR sweeps and 1.4M object bounding boxes in 40k keyframes. 23 object classes with accurate 3D bounding boxes at 2Hz over the entire dataset are annotated. Additionally, object-level attributes such as visibility, activity and pose are annotated.

\textbf{NOCS} \cite{wang2019normalized} is now the most famous dataset for category-level object pose detection. It consists of the synthetic CAMERA275 dataset and the real-world  REAL25 dataset. Six categories are involved in NOCS. For CAMERA275, 300K composited images are rendered, among which 25K are set aside for validation.  A total of 184 instances selected from ShapeNetCore  \cite{chang2015shapenet}, and 31 widely varying indoor scenes (backgrounds) are used for rendering.  For REAL25,  8K RGBD frames (4300 for training, 950 for validation, and 2750 for testing) of 18 different real scenes (7 for training, 5 for validation, and 6 for testing) are captured.  In NOCS, the resolution of each image is $640 \times 480$. For each training and testing subset, 6 categories and 3 unique instances per category are used. For the validation set, six categories with one unique instance per category are used. More than 5 object instances are placed in each scene to simulate real-world clutter. CAD models of  instances are available for CAMERA275 and REAL25. All test images are annotated with 9Dof parameters w.r.t pose and size. The fly in the ointment is that object poses and sizes in training datasets are roughly obtained through registering, which are not as accurate as test annotations.

\textbf{Objectron} \cite{ahmadyan2020objectron} dataset is a collection of short object-centric video clips, accompanied by AR session metadata, including camera poses, sparse point clouds, and characterization of planar surfaces in the surrounding environment. In each video, the camera moves around the object, capturing it from different views. This dataset also contains manually annotated 3D bounding boxes for each object, which describe the object’s position, orientation, and dimensions. Objectron consists of 15K annotated video clips supplemented with over 4M annotated images belonging to categories of bikes, books, bottles, cameras, cereal boxes, chairs, cups, laptops, and shoes. The original  resolution of each image is $1920 \times 1440$. In addition, this dataset is collected from 10 countries across five continents to ensure geo-diversity. Owing to the richness of this dataset, it is very suitable for evaluating the performance of category-level monocular object pose detection/tracking methods, especially when only RGB data is used. The disadvantage is that most images contain only one object, and it occupies most of the image area, which is not conducive to training several object pose detection or tracking models for robot grasping. However, for AR scenarios, this dataset is simply a boon for the industry.

\section{Instance-Level Monocular Object Pose Detection}
\label{secinstance}

Instance-level monocular object pose detection aims to detect the object of interest and estimate its 6Dof pose (i.e., 3D rotation $\mathcal{R} \in SO(3)$ and 3D translation $\mathcal{T} \in R^3$) relative to a canonical frame. Particularly, the CAD model of the interested object is available during training and testing. According to different input data formats, instance-level monocular object pose detection methods are classified into \textbf{RGB-based methods} and \textbf{(RGB)D-based methods}.

\begin{figure*} 
	\centering 
	\subfigure[RGB-based methods]{\includegraphics[width=2in]{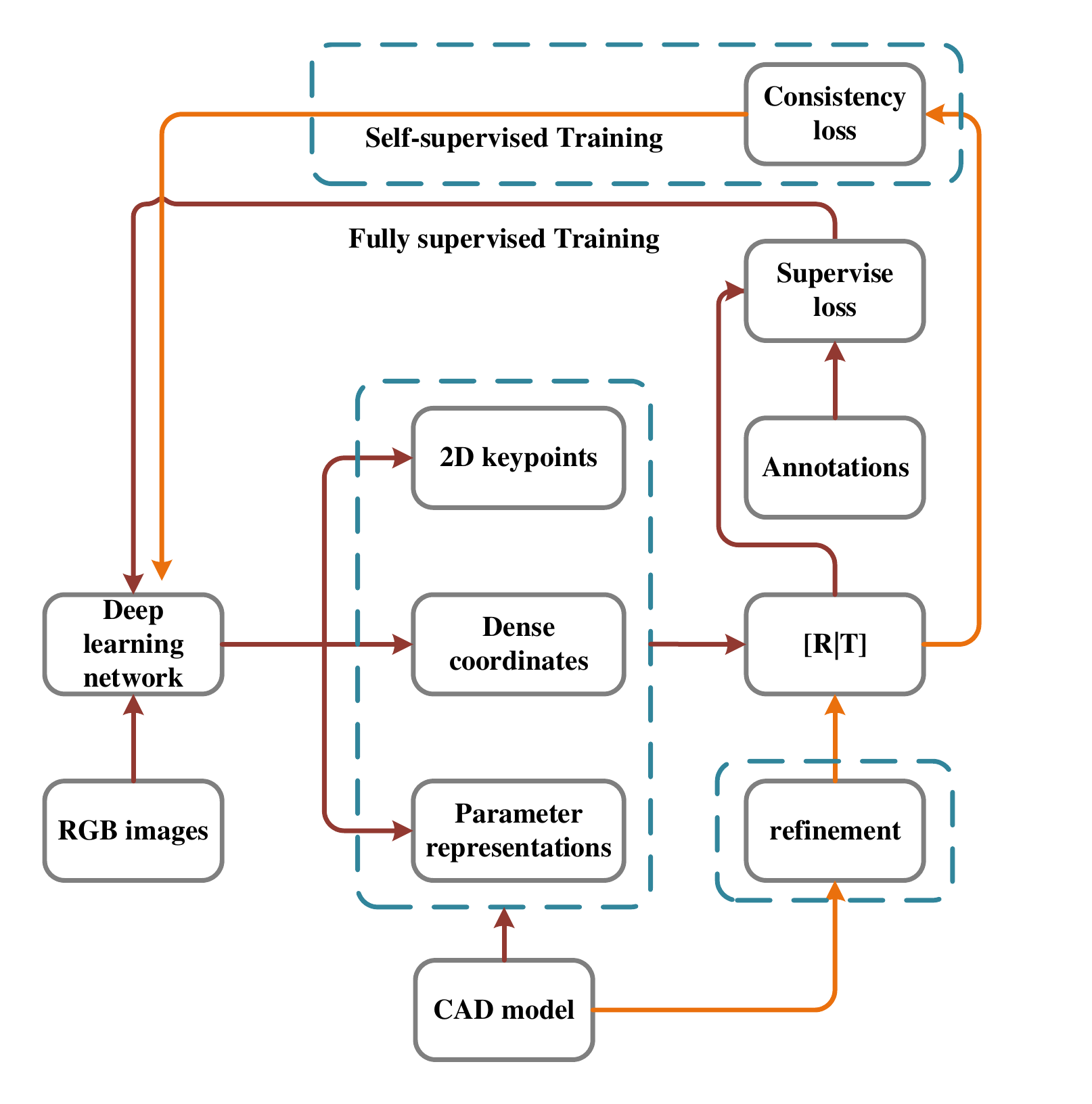}\label{instance_rgb}} 
	\subfigure[(RGB)D-based methods]{\includegraphics[width=2in]{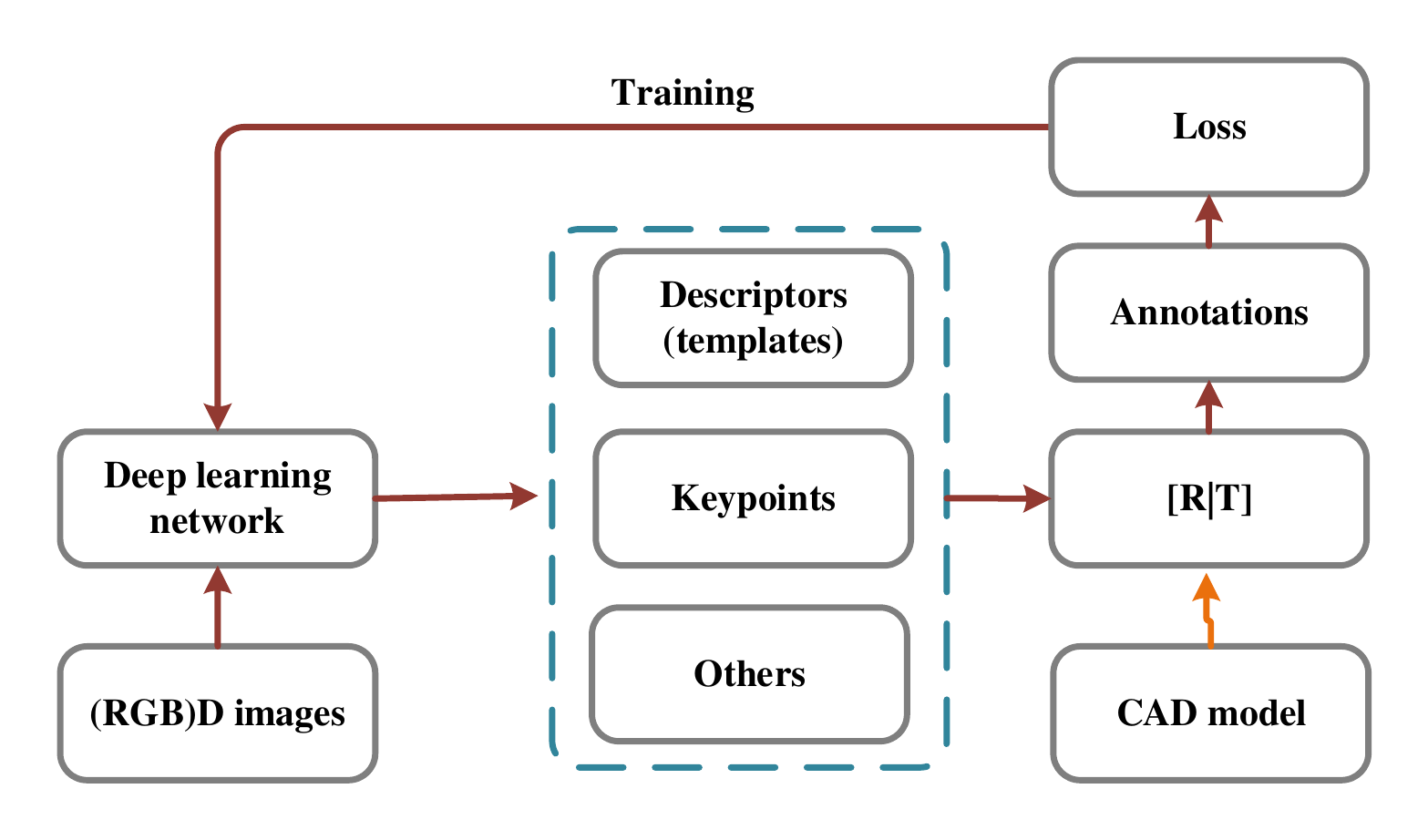}\label{instance_rgbd}} 
	\vspace{-0.2in} 
	\caption{ Overall schematic representation of instance level monocular object pose detection methods} 
	\vspace{-0.3in} 
\end{figure*} 

\subsection{RGB-based Methods}
The most direct way to estimate the 6Dof pose is to let the deep learning model predict pose-related parameters. However, directly estimating the 6Dof pose from a single RGB image is an ill-posed problem and faces many challenges. Owing to the existence of CAD models,  establishing 2D-3D correspondences between the input image and the object model can help ease the task. According to the above observation, Figure \ref{instance_rgb} provides an overall schematic representation of RGB-based instance-level monocular object pose detection methods. In general, deep learning-based methods are divided into five major classes: direct methods, keypoint-based methods, dense coordinate-based methods, refinement-based methods, and self-supervised methods.

\textbf{Direct methods.} One of the most intuitive ways to predict the 6Dof pose is to treat object pose estimation as a regression or classification task and directly predict object pose-related parameter presentations (for example, Euler angles or quaternions for rotations) from input images. For instance, Xiang et al. \cite{xiang2017posecnn} introduce a novel deep learning model for end-to-end 6Dof pose estimation named PoseCNN, which decouples object pose estimation task into several different components. For translation estimation, PoseCNN first localizes the 2D object center in the image and estimates the depth w.r.t the camera. The 3D translation can be recovered according to the projection equation. The 3D rotation is estimated by regressing a quaternion representation. Both components have utilized geometric priors to simplify the task. SSD-6D \cite{kehl2017ssd} is another direct prediction pipeline. It extends the 2D SSD \cite{liu2016ssd} architecture to 6Dof pose estimation. Instead of regressing rotation parameters continuously, they discretize the rotation space into classifiable viewpoint bins and treat rotation estimation as a classification problem. Recently, Trabelsi et al. \cite{trabelsi2021pose} put forward a pose proposal network for 6Dof object pose estimation and presented a CNN-based encoder/multi-decoder module. The multidecoder decouples rotation estimation, translation estimation, and confidence estimation into different decoders. Hence, specific features can be learned for each subtask. 

Though intuitive, all the above methods are highly dependent on time-consuming pose refinement operation to improve performance. By contrast, the recent work EfficientPose \cite{bukschat2020efficientpose} introduces a direct pose estimation framework based on the state-of-the-art EfficientDet \cite{tan2020efficientdet} architecture. By utilizing the novel 6D augmentation strategy, EfficientPose achieves superior performance without refinement. However, as it is an ill-posed problem, direct methods always perform poorly in natural scene generalization. Currently, the better, more popular choice is to establish 2D-3D correspondences and solve them to predict pose-related parameters, which will be introduced in the next several subsections. 

Recently, another branch of works \cite{hu2020single, chen2020end, wang2021gdr} adopts such ideas in direct predicting methods. More specifically, they try to modify the indirect methods to direct methods by utilizing neural networks to establish 2D-3D correspondences directly and simulating the Perspective-n-Point (PnP) \cite{lepetit2009epnp} algorithm using deep learning networks. In this way, object pose can be directly regressed by the correspondence-extraction network combined with the PnP network. For example, GDR-Net \cite{wang2021gdr} first regresses geometric features using a ResNet \cite{he2016deep} backbone to build 2D-3D correspondences. Then, the 2D-3D correspondences are fed to a Patch-PnP solver consisting of CNNs and fully connected layers to predict object pose. The whole framework of GDR-Net can be trained end-to-end. Though requiring 2D-3D correspondences, the object pose can be directly predicted by a network. Thus, these works are also regarded as direct methods in this study. 

The advantages of direct methods are that they are often more lightweight and easy to be trained. However, their performance is usually poorer than that of other methods because they heavily rely on the deep networks only to recover the full 6Dof parameters.

\textbf{Keypoint-based methods}. As mentioned before, compared with directly predicting pose-related parameters, building 2D-3D correspondences for object pose detection is more accurate. Among current correspondence-based methods,  keypoint-based methods first use CNNs to detect 2D keypoints in the image and then solve a PnP problem \cite{ lepetit2009epnp}  for pose estimation, achieve remarkable performance. As the pioneer, BB8 \cite{ rad2017bb8} first uses a deep network to segment out the object coarsely and then uses another network to predict 2D projections of the 3D bounding box corners to build 2D-3D correspondences. Based on such 2D-3D correspondences, the 6Dof pose can be estimated by solving the PnP problem. Owing to its multi-stage pipeline, BB8 cannot be trained end-to-end and is time consuming in inference.  Similar to BB8, Tekin et al. \cite{ tekin2018real} adopt the idea of YOLO  \cite{redmon2016you}  framework for 2D detection and extend it for 6Dof object pose detection, namely YOLO-6D. The key component of YOLO-6D is a single-shot deep network that takes the image as input and directly detects the 2D projections of 3D bounding box corners, followed by a PnP solver to estimate object pose. This two-stage pipeline yields a fast, accurate 6Dof pose prediction without requiring any post processing. 

After that, this kind of two-stage pipeline method is adopted by many works \cite{pavlakos20176,zhao2018estimating,oberweger2018making,hu2019segmentation}, addressing different challenges such as occlusion and truncation. For instance, Oberweger et al. \cite{oberweger2018making} propose predicting the 2D projections of 3D keypoints in the form of 2D heatmaps. Compared with directly regressing the coordinates of 2D keypoints, localizing keypoints from heatmaps addresses the issue of occlusion to some extent. However, given that heatmaps are fix in size, handling truncated objects is challenging because several of their keypoints may be outside the image. To solve this problem, PVNet \cite{peng2019pvnet} adopts the strategy of voting-based keypoint localization. Specifically, it first trains a CNN to regress pixel-wise vectors pointing to the keypoints, that is, vector field, and then uses the vectors belonging to the target object pixels to vote for keypoint locations. Owing to this vector-field representation, the occluded or truncated keypoints can be robustly recovered from the visible parts, enabling PVNet to yield good performance under severe occlusion or truncation. 

Inspired by PVNet, several works also perform a pixel-wise voting scheme for keypoint localization to improve performance. For instance, Yu et al. \cite{yu20206dof} propose an effective loss on top of PVNet to achieve more accurate vector-field estimation by incorporating the distances between pixels and keypoints into the objective. Then, HybridPose \cite{song2020hybridpose} extends unitary intermediate representation to a hybrid one, which includes keypoints, edge vectors, and symmetry correspondences. This hybrid representation exploits diverse features to enable accurate pose prediction. However, regressing more representations also causes more computational cost and limits inference speed.

Existing state-of-the-art keypoints-based methods are accurate and robust toward occlusion and truncation, which is a remarkable advantage. However, they require predefining keypoints for each object, which is quite a labor cost. Moreover, the pose detection results would be highly affected by the keypoints detection results. Therefore, in case of failure, analyzing the reason is difficult.

\textbf{Dense coordinate-based methods}. Beyond keypoint-based methods, another kind of correspondence-based method is dense coordinate-based methods, which formulate the 6Dof object pose estimation task as building dense 2D-3D correspondences, followed by a PnP solver to recover the object pose. The dense 2D-3D correspondences are obtained by predicting the 3D object coordinate of each object pixel or by predicting dense UV maps. Before deep learning becomes popular, early works usually use random forest for predicting object coordinates \cite{krull20146, brachmann2014learning, nigam2018detect}. Brachmann et al. \cite{brachmann2016uncertainty} extend the standard random forest to an auto context regression framework, which iteratively reduces the uncertainty of the predicted object coordinate.

However, the performances of these early works are poor because random forest can only learn limited features in simple scenarios. To handle more complicated conditions, recent studies introduce deep learning networks to predict 3D coordinates of object pixels. For instance, CDPN \cite{li2019cdpn} is a pioneer work in adopting deep learning models for dense coordinates prediction. In CDPN, the predicted 3D object coordinates are used to build dense correspondences between the 2D image and the 3D model, which has shown robustness toward occlusion and clutter. However, CDPN neglects the importance of handling symmetries.  To estimate the 6Dof pose of symmetric objects accurately, Pix2Pose \cite{park2019pix2pose} builds a novel loss function, namely transformer loss, which can transform the predicted 3D coordinate of each pixel to its closest symmetric pose. 

The above methods directly regress dense 3D object coordinates, which is also an ill-posed problem and may face challenges for continuously predicting. Instead of directly regressing 3D coordinates, DPOD \cite{zakharov2019dpod} builds 2D-3D correspondences by employing UV maps. Given a pixel color, its corresponding position on the 3D model surface can be estimated based on the correspondence map, thus providing a relation between image pixels and 3D model vertices. Moreover, regressing UV maps turns out to be a much easier task for the network. Whether to predict coordinates or UV maps, the above methods make predictions for all visible points on the object, which is inefficient and difficult to learn. In EPOS \cite{hodan2020epos}, the target object is represented by compact surface fragments, and correspondences between densely sampled pixels and fragments are predicted using an encoder-decoder network, which is much more efficient and geometrically more reasonable. Moreover, by doing so,  challenging cases such as objects with global or partial symmetries can be solved to a certain degree. 

In summary, dense coordinate-based methods are robust to heavy occlusions and symmetries. However, regressing object coordinates remains more difficult than predicting sparse keypoints due to the larger continuously searching space.

\textbf{Refinement-based methods}.  As mentioned previously, several methods require a refinement step to improve their performances. Here, these methods are called refinement-based methods, which estimate the object pose by aligning synthetic object renderings with real observed images. This kind of method may overlap with several methods that have been mentioned above, but they are also introduced separately here because they are usually the key to improving the prediction performance.

Given an initial pose estimation, DeepIM \cite{li2018deepim} exploits a deep learning-based pose refinement network to refine the initial pose iteratively by minimizing the differences between the observed image and the rendered image under the current pose. The refined pose is then used as the initial pose for the next iteration, until the refined pose converges or the number of iterations reaches a threshold. Manhardt et al. \cite{manhardt2018deep} also adopt the iterative matching pipeline and propose a new visual loss that drives the pose update by aligning object contours. Moreover, DPOD \cite{zakharov2019dpod} develops a standalone pose refinement network including three separate output heads for regressing rotation and translation. In the work of Trabelsi et al. \cite{trabelsi2021pose}, the attention mechanism is integrated into a deep network for refining the initial pose by highlighting important parts. The above methods focus on comparing the rendered image and the observed image, ignoring the importance of realistic rendering. Recently, Yen-Chen et al. \cite{yen2020inerf} propose a framework iNeRF to invert an optimized neural radiance field for rendering and has achieved promising results.

These refinement networks have already been integrated into several instance-level object pose detection models to produce more accurate results. For example, PoseCNN\cite{xiang2017posecnn} is combined with DeepIM\cite{li2018deepim}, which have shown great advantages by adopting refinement as a post processing step. However, the time consumption of refinement-based methods depends heavily on the number of iterations and the used renderer, which becomes a bottleneck for their wide application.

\begin{table}
	\centering
	\resizebox{\textwidth}{30mm}{
		\begin{tabular}{ r | c| c| c| c| c| c  }  
			\hline 	
			\multirow{2}*{Methods}& \multirow{2}*{Types} &\multirow{2}*{Input} &\multirow{2}*{Pose refinement} &2D projection  &ADD(-S)  &5 \degree 5 cm \\
			
			& &  & & metric&metric&metric\\
			\hline 
			Brachmann et al.(2016) \cite{brachmann2016uncertainty}& Dense&RGB &Yes& 73.7  &50.2 & 40.6 \\
			
			BB8(2017) \cite{rad2017bb8}&Keypoints &RGB &Yes& 89.3  &62.7 & 69.0 \\
			SSD-6D(2017) \cite{kehl2017ssd}& Direct&RGB &Yes &- &76.3 &-  \\
			
			PoseCNN(2017) \cite{xiang2017posecnn}&Direct &RGB &No & 70.2  &62.7 & 19.4 \\
			DeepIM(2018) \cite{li2018deepim} & Refinement&RGB &Yes & 97.5  &88.6 & 85.2 \\
			
			YOLO-6D(2018) \cite{tekin2018real}&Keypoints&RGB&No&90.37 &55.95 &-\\
			Sundermeyer et al.\cite{sundermeyer2018implicit}&Self-supervised&RGB&No&- &28.65 &-\\
			
			Pix2Pose(2019) \cite{park2019pix2pose}& Dense&RGB &No &- &72.4 &-  \\	
			PVNet(2019) \cite{peng2019pvnet}& Keypoints&RGB &No &99.00 &86.27 &-  \\	
			CDPN(2019) \cite{li2019cdpn}&Dense &RGB &No &98.10 &89.86 &94.31  \\	
			DPOD(2020) \cite{zakharov2019dpod}& Dense&RGB &Yes &- &95.15 &- \\
			Yu et al. (2020) \cite{yu20206dof}&Keypoints &RGB &No&99.40 &91.50 &-  \\
			
			HybridPose(2020) \cite{song2020hybridpose}& Keypoints&RGB &Yes &- &91.30 &- \\
			Self6D(2020) \cite{wang2020self6d}&Self-supervised &RGB &No &- &58.9 &- \\
			iNeRF(2020) \cite{yen2020inerf}& Self-supervised &RGB &Yes &- &79.2 &- \\
			Sock et al. \cite{sock2020introducing}&Self-supervised &RGB &No &- &60.6 &- \\
			Li et al. (2020) \cite{li2020robust}&Self-supervised &RGB &No &- &80.4 &-  \\		
			
			EfficientPose(2021) \cite{bukschat2020efficientpose}& Direct&RGB &No &- &97.35 &- \\
			Trabelsi et al.(2021) \cite{trabelsi2021pose}&Direct &RGB &Yes & 99.19&93.87 &- \\
			GDR-Net(2021) \cite{wang2021gdr}&Direct &RGB &No &- &93.7 &-  \\

			DSC-PoseNet(2021) \cite{yang2021dsc}&Self-supervised &RGB &No &- &58.6 &-  \\	
			\hline 
			
	\end{tabular}}
	\caption{Perfomance of instance level RGB-based methods on Linemod dataset.}
	\label{linemod}
	\vspace{-0.3in} 
\end{table}

\begin{table*}
	\centering
	\resizebox{\textwidth}{25mm}{
		\begin{tabular}{ r |c| c |c| c| c |c  }  
			\hline 	
			\multirow{2}*{Methods}&\multirow{2}*{Types}&\multirow{2}*{Input} &\multirow{2}*{Pose refinement} &2D projection  &ADD(-S)  &5 \degree 5 cm \\
			
			& &  & & metric &metric &metric \\
			\hline 
			
			PoseCNN(2017) \cite{xiang2017posecnn}& Direct&RGB &No& 17.2  &24.9 &- \\
			DeepIM(2018) \cite{li2018deepim}& Refinement &RGB &Yes & 56.6  &55.5 & 30.9 \\
			
			YOLO-6D(2018) \cite{tekin2018real}& Keypoints&RGB&No&6.16 &6.42  &-  \\
			Oberweger et al. (2018) \cite{oberweger2018making}&Keypoints &RGB &No&60.9 &30.4 &- \\
			Pix2Pose(2019) \cite{park2019pix2pose}&Dense &RGB &No &- &32.0 &- \\	
			
			SegDriven(2019) \cite{hu2019segmentation}&Keypoints &RGB &No &44.9 &27.0 &- \\	
			PVNet(2019) \cite{peng2019pvnet}& Keypoints&RGB &No &61.06 &40.77 &- \\	
			DPOD(2020) \cite{zakharov2019dpod}&Dense &RGB &Yes &- &47.25 &- \\
			
			Yu et al. (2020) \cite{yu20206dof}&Keypoints &RGB &No&-  &43.52 &-  \\
			Single-stage(2020) \cite{hu2020single}&Direct &RGB &No&62.3 &43.3 &- \\
			HybridPose(2020) \cite{song2020hybridpose}&Keypoints &RGB &Yes &- &47.5 &- \\
			Self6D(2020) \cite{wang2020self6d}&Self-supervised &RGB &No &- &32.1 &- \\
			Sock et al. \cite{sock2020introducing}&Self-supervised &RGB &No &- &22.8 &- \\
			Li et al. (2020) \cite{li2020robust}&Self-supervised &RGB &No &- &42.3 &- \\
			
			GDR-Net(2021) \cite{wang2021gdr}& Direct&RGB &No &- &62.2 &- \\
			Trabelsi et al.(2021) \cite{trabelsi2021pose}&Direct &RGB &Yes & 65.46&58.37 &- \\

			DSC-PoseNet(2021) \cite{yang2021dsc}&Self-supervised &RGB &No &- &24.8 &- \\
			\hline 
			
	\end{tabular}}
	\caption{Perfomance of instance level RGB-based methods on the Occlusion Linemod dataset.}
	\label{occ_linemod}
	\vspace{-0.25in} 
\end{table*}

\begin{table*}
	\centering
	\resizebox{\textwidth}{15mm}{
		\begin{tabular}{r |c| c| c |c |c| c  }  
			\hline 	
			\multirow{2}*{Methods}&\multirow{2}*{Types}&\multirow{2}*{Input} &\multirow{2}*{Pose refinement} &2D projection &ADD(-S)  &ADD(-S) AUC\\
			
			&&  & & metric&metric &metric\\
			\hline 
			
			PoseCNN(2017) \cite{xiang2017posecnn}&Direct&RGB &No &3.7 &21.3 &61.3\\
			DeepIM(2018) \cite{li2018deepim} &Refinement &RGB &Yes & - & -  &81.9\\
			Oberweger et al. (2018)\cite{oberweger2018making}&Keypoints&RGB &No&39.4 &- &72.8 \\
			SegDriven(2019) \cite{hu2019segmentation}&Keypoints&RGB &No &30.8 &39.0 &- \\	
			PVNet(2019) \cite{peng2019pvnet}& Keypoints&RGB &No &47.4 &- &73.4 \\	
			
			Single-stage(2020) \cite{hu2020single}&Direct&RGB &No&48.7 & 53.9 &- \\
			GDR-Net(2021) \cite{wang2021gdr}&Direct&RGB &No &-  &60.1 &84.4 \\
			Li et al. (2020) \cite{li2020robust}&Self-supervised&RGB &No &15.6 &- &50.5 \\
			
			\hline 
			
	\end{tabular}}
	\caption{Perfomance of instance level RGB-based methods on the YCB-Video dataset.}
	\label{ycb}
	\vspace{-0.4in} 
\end{table*}

\textbf{Self-supervised methods}. In the 6Dof object pose detection task, current deep learning models highly rely on training on annotated real-world data, which are difficult to obtain. An intuitive way is to use costless synthetic data for training. Nevertheless, many works has proven that models simply trained on synthetic data demonstrate poor generalization ability toward real-world scenarios due to the vast domain gap between synthetic data and natural data. 

Therefore, domain randomization (DR) \cite{tobin2017domain} is proposed and used \cite{sundermeyer2018implicit}. Its core idea is to generate enormous images by sampling random object poses and placing the model on randomly selected background images. In this way, the real-world domain is only a subset of the generated domain. The model can learn as much 6Dof pose-related features as possible by utilizing these synthetic images. In addition, many methods \cite{movshovitz2016useful} attempt to generate more realistic renderings \cite{movshovitz2016useful, wen2020se,wang2019normalized} or use data augmentation strategies \cite{peng2019pvnet, bukschat2020efficientpose}. 

However, these schemes cannot address challenges such severe occlusions, and the performance on real data is still far from satisfactory.
In this context, the idea of self-supervised learning is introduced. For instance, Deng et al. \cite{deng2020self} propose a novel robot system to label real-world images with accurate 6D object poses. By interacting with objects in the environment, the system can generate new data for self-supervised learning and improve its pose estimation results in a life-long learning fashion. Recently, Self6D \cite{wang2020self6d} leverages the advances in neural rendering \cite{chen2019learning} and proposes a self-supervised 6Dof pose estimation solution. Self6D first trains the network on synthetic data in a fully supervised manner and then fine-tunes on unlabeled real RGB data in a self-supervised manner by seeking a visually and geometrically optimal alignment between real images and rendered images. 

Similarly, Sock et al. \cite{sock2020introducing} propose a two-stage 6Dof object pose estimator framework. In their work, the first stage enforces the pose consistency between rendered predictions and real input images, narrowing the gap between these two domains. The second stage fine-tunes the previously trained model by enforcing the photometric consistency between pairs of different object views, where one image is warped and aligned to match the view of the other, thus enabling their comparison. Additionally, the above introduced iNeRF \cite{yen2020inerf} has no need to use any real annotated data for training. Therefore, it also belongs to this line of work.

Beyond leveraging image-level consistency for self-supervised learning, inspired by recent keypoint-based methods \cite{peng2019pvnet, song2020hybridpose}, DSC-PoseNet \cite{yang2021dsc} develops a weakly supervised and a self-supervised learning-based pose estimation framework that enforces dual-scale keypoint consistency without using pose annotations. Recently, Li et al. \cite{li2020robust} use the consistency between an input image and its DR augmented counterpart to achieve self-supervised learning. As a result, this self-supervised framework can robustly and accurately estimate the 6Dof pose in challenging conditions and ranks the current best self-supervised based instance-level monocular object pose detection model.

Briefly, self-supervised methods are advanced to save annotation costs, but their performance is still far from satisfactory. Further improvements should be made to reduce domain gap between self-supervised training and full-supervised training.

Thus far, most of the recent state-of-the-art RGB-based instance-level monocular object pose detection methods from several different perspectives have been introduced, and full performance comparisons of most of these above-mentioned methods have been provided in Table \ref{linemod}, \ref{occ_linemod}, and \ref{ycb} on Linemod dataset, Occlusion Linemod dataset, and YCB-Video dataset, respectively. 

\begin{table*}
	\centering
	\resizebox{\textwidth}{25mm}{
		\begin{tabular}{r |c|c|c|c|c|c }
			
			\hline
			Method & Type & Year & Input & Pose Refinement & LINEMOD\cite{hinterstoisser2012model} & YCB video\cite{xiang2017posecnn}\\
			
			\hline
			
			DenseFusion\cite{wang2019densefusion} & Other & 2019 & RGB-D & Yes & - & 96.8 \\
			
			MoreFusion\cite{wada2020morefusion} & Other & 2020 & RGB-D & Yes & - & 95.7 \\
			
			PVN3D\cite{he2020pvn3d} & Keypoints & 2020 & RGB-D & No & 95.5 & - \\
			
			PVN3D\cite{he2020pvn3d}+ICP & keypoints & 2020 & RGB-D & Yes & 96.1 & - \\
			
			Gao et al.\cite{gao20206d} & Other & 2020 & Depth & Yes & - & 82.7 \\
			
			Gao et al.\cite{gao20206d}+ICP & Other & 2020 & Depth & Yes & - & 76.0 \\
			
			G2L-Net\cite{chen2020g2l} & Other & 2020 & RGB-D  & No & 98.7 & 92.4  \\
			
			CloudAAE\cite{gao2021cloudaae} & Other & 2021 & Depth & No & 82.1 & -  \\

			CloudAAE\cite{gao2021cloudaae}+ICP & Other & 2021 & Depth & Yes & 92.5 & 93.5 \\

			RCVPose\cite{wu2021vote} & Keypoints & 2021 & RGB-D & No & 99.4 & 95.2 \\

			RCVPose\cite{wu2021vote}+ICP & Keypoints & 2021 & RGB-D & Yes & 99.7 & 95.9 \\
			
			FFB6D\cite{he2021ffb6d} & Keypoints & 2021 & RGB-D & No & 99.7 & 97.0 \\
			
			FFB6D\cite{he2021ffb6d}+ICP & Keypoints & 2021 & RGB-D & No & - & 96.6 \\
			
			\hline
	\end{tabular}}
	\caption{Performance comparison of (RGB)-D based  instance level monocular 6Dof pose detection methods. The metric is ADD(-S).}
	\label{tab:rbgb_methods}
	\vspace{-0.4in} 
\end{table*}

\subsection{(RGB)D-based Methods}
RGB images lack depth information, making the 6Dof object pose detection task an ill-posed problem. However,  the development of monocular RGBD cameras motivates the (RGB)D-based 6Dof pose estimation methods. (RGB)D-based methods take RGBD images or depth masks as input and predict object pose by fully utilizing of the power of the point cloud presentation. In general, the (RGB)D-based methods can be classified into retrieval-based methods, keypoint-based methods, and other deep learning-based methods. Figure \ref*{instance_rgbd} presents an overall schematic representation of (RGB)D-based instance-level monocular object pose detection methods.

\textbf{Retrieval-based methods}. A direct solution for predicting object pose is to traverse all possibilities. Using a CAD model, a dataset of synthetic (RGB)D images covering all possible object poses can first be generated. Then, we can retrieve the most similar image of the target object from the dataset can be retrieved to determine the object pose given an observed RGBD image or depth mask. The retrieval can be achieved by comparing image descriptors or matching templates. In this study, this kind of method is called retrieval-based method. Traditional methods \cite{huttenlocher1993comparing,steger2001similarity,hinterstoisser2011gradient, hinterstoisser2011multimodal, hinterstoisser2012model} achieve retrieving or template matching by using handle-craft features, hence possess limited perception power. Motivated by traditional retrieval-based methods, to the best of our knowledge,  Kehl et al. \cite{kehl2016deep} is the first to propose to train a convolutional auto encoder \cite{masci2011stacked} from scratch using  RGBD patches to learn descriptors for retrieval. When new RGBD images come, the target object poses can be predicted by matching object descriptors with these descriptors in the codebook. 

Though considerable improvements have been obtained compared with traditional methods, the work of Kehl et al. \cite{kehl2016deep} faces challenges in handling occlusion and textureless objects. Therefore, Park et al. \cite{park2019multi} put forward multitask template matching (MTTM). MTTM only uses depth masks as input to find the closest template of the target object. In this pipeline, object segmentation and pose transformation are performed to remove irrelevant points. By eliminating irrelevant points, MTTM shows great robustness in pose estimation.
MTTM can be generalized to the category level because object segmentation and pose transformation are instance independent. However, the performance of MTTM  decreases largely without using CAD models. 

In contrast to the above methods that directly learn descriptors or templates relying on the network architecture, several methods introduce triplet comparison in their pipelines to learn more powerful descriptors. For example,  Alexander et al. \cite{wohlhart2015learning} propose a deep learning-based descriptor predicting method using triplet loss, which successfully untangles different objects and views into clusters. Then, based on this work \cite{wohlhart2015learning}, Balntas et al. \cite{balntas2017pose} further develop a new descriptor representation that enforces a direct relationship between the learned features and pose label differences, which helps better represent object pose in latent space. Next, Zakharov et al. \cite{zakharov20173d} draw an angular difference function and set a constant angle threshold into triplet loss as a dynamic margin to speed up training and obtain better performance. 

In summary, retrieval-based methods can achieve robust performance in most cases, but they need to discretize the rotation space for defining codebooks. On the one hand, it would result in rough object pose predictions when the discrete interval is large. On the other hand, it would lead to a substantial computational cost when the discrete gap is small.

\textbf{Keypoints-based methods} Similar to RGB-based methods, a line of works attempts to predict object pose by extracting object keypoints in RGB(D)-based methods. This kind of method always constructs 3D-3D correspondence to solve object pose utilizing predicted keypoints. PVN3D \cite{he2020pvn3d} may be the first deep learning work that successfully estimates object pose by extracting 3D keypoints. A deep Hough voting network is designed to detect 3D keypoints of objects and then estimate the 6D pose parameters within a least-squares fitting manner.  Then, PointPoseNet \cite{chen2020pointposenet} combines the advantage of color and geometry information to improve PVN3D. Specifically, PointPoseNet simultaneously predicts segmentation masks and regresses
point-wise unit vectors pointing to the 3D keypoints. Then unit vectors of segmented points are used to generate the best pose hypothesis with the help of geometry constraints. PVN3D and PointPoseNet opt to use keypoints selected by Farthest Point Sampling (FPS) algorithm.

By contrast, full flow bidirectional fusion network designed for 6D pose estimation) \cite{he2021ffb6d} comes up with a novel keypoint selection algorithm called SIFT-FPS as replacement of FPS to leverage texture and geometry information fully. SIFT-FPS formulates two steps. In the first step, SIFT \cite{lowe2004distinctive} features are extracted to extract 2D keypoints. In the second step, 2D keypoints are transformed into 3D and further selected by FPS. In this way, the selected keypoints could be more evenly distributed on the object surface and be more texture distinguishable.

More recently,  Wu et al. \cite{wu2021vote} argue that previous vector or offset schemes are sensitive to disperse keypoints. They propose model RCV-Pose. In  RCV-Pose, the radial voting scheme is introduced, where for each object point, a deep learning model is used to learn several spheres. Each keypoint may locate on the surface of a sphere. Therefore, during inference,  the algorithm only needs to use the surface of these spheres to vote for a 3D accumulator space, whose peaks indicate keypoint locations. Compared with previous methods, RCV-Pose only needs to predict three keypoints, which is much more efficient. 

In summary, the keypoints-based methods are robust to noise and can achieve relatively accurate detection results. The limitation is they are sensitive to the way keypoints are defined. Moreover, they often require iterative optimization with high time complexity.

\textbf{Other methods}. In addition to the above methods, several works have shown great interest in designing models that satisfy several special requirements beyond pursuing accuracy. 

First, several methods have attempted to solve problems caused by complex scenarios such as occlusion and clutter. For instance,  Alexander et al. \cite{krull2015learning} propose an approach that \emph{learns to compare} the observed RGBD image and the rendered image to handle occlusion and complicated sensor noise better by describing the posterior density of a particular object pose with a convolutional neural network (CNN) that compares observed images and rendered images. Jafari et al. \cite{jafari2018ipose} present a three-step decomposition approach with the help of instance segmentation and dense coordinate regression to handle occlusion, known as the first deep learning-based accurate pose estimator for partly occluded objects. Then, Wang et al. \cite{wang2019densefusion} propose DenseFusion, which fully leverages color and depth information, leading to robustness in heavy occlusion and changing lighting conditions. Furthermore, MoreFusion \cite{wada2020morefusion} performs pose prediction with surrounding spatial awareness and joint multiple object pose optimization, which greatly promotes consistency and accuracy of pose estimation in cluttered scenes with heavy occlusion and contracting objects. 

Second, besides handling complex scenarios, recent studies have shown interest in designing more lightweight network architectures to prompt real-time performance. The work of Gao et al. \cite{gao20206d} is known as the first point cloud-only based deep learning 6Dof pose estimation method. The network architecture proposed by Gao et al. \cite{gao20206d} is much more lightweight than methods that take RGB data and depth data as input because only point cloud is required to be processed. Then, Chen et al. \cite{chen2020detecting} use segmented object point cloud patches from the target object's CAD model as input and form a pose prediction model without the need for fine-tuning, thus saving computational cost. Additionally, G2L-Net \cite{chen2020g2l} decouples the prediction pipeline into global localization, translation localization, and rotation localization and estimates the object pose in a coarse-to-fine manner without using a CAD model, achieving real-time performance.

Third, in addition to considering running time,  Gao et al. \cite{gao2021cloudaae} improve the generalization ability of models trained on synthetic data. They argue that the domain gap between real and synthetic is considerably smaller and easier to fill for depth information. Therefore, they present an end-to-end 6Dof pose prediction architecture called CloudAEE. It utilizes an augmented autoencoder and a lightweight depth data synthesis pipeline to cooperatively improve generalization ability cooperatively.

Thus far, most of the recent state-of-the-art (RGB)D-based instance-level monocular object pose detection methods have been introduced in this work, and a full performance comparison of keypoints-based methods and other methods is presented in Table \ref{tab:rbgb_methods}. The metric is ADD(-S). Retrieval-based methods do not use commonly used datasets or unified experimental settings; thus, they are not quantitatively compared in this survey. 

\section{Category-Level Monocular Object Pose Detection}
\label{seccatefory}
In this section, category-level monocular object pose detection methods are introduced. According to whether the prediction focuses on 1Dof rotation or 3Dof rotation, related methods are classified into \textbf{Category-Level Monocular 3D Object Detection} and \textbf{Category-Level Monocular 6D Pose Detection}.

\begin{figure*} 
	\centering 
	\subfigure[3D object dectection methods]{\includegraphics[width=2in]{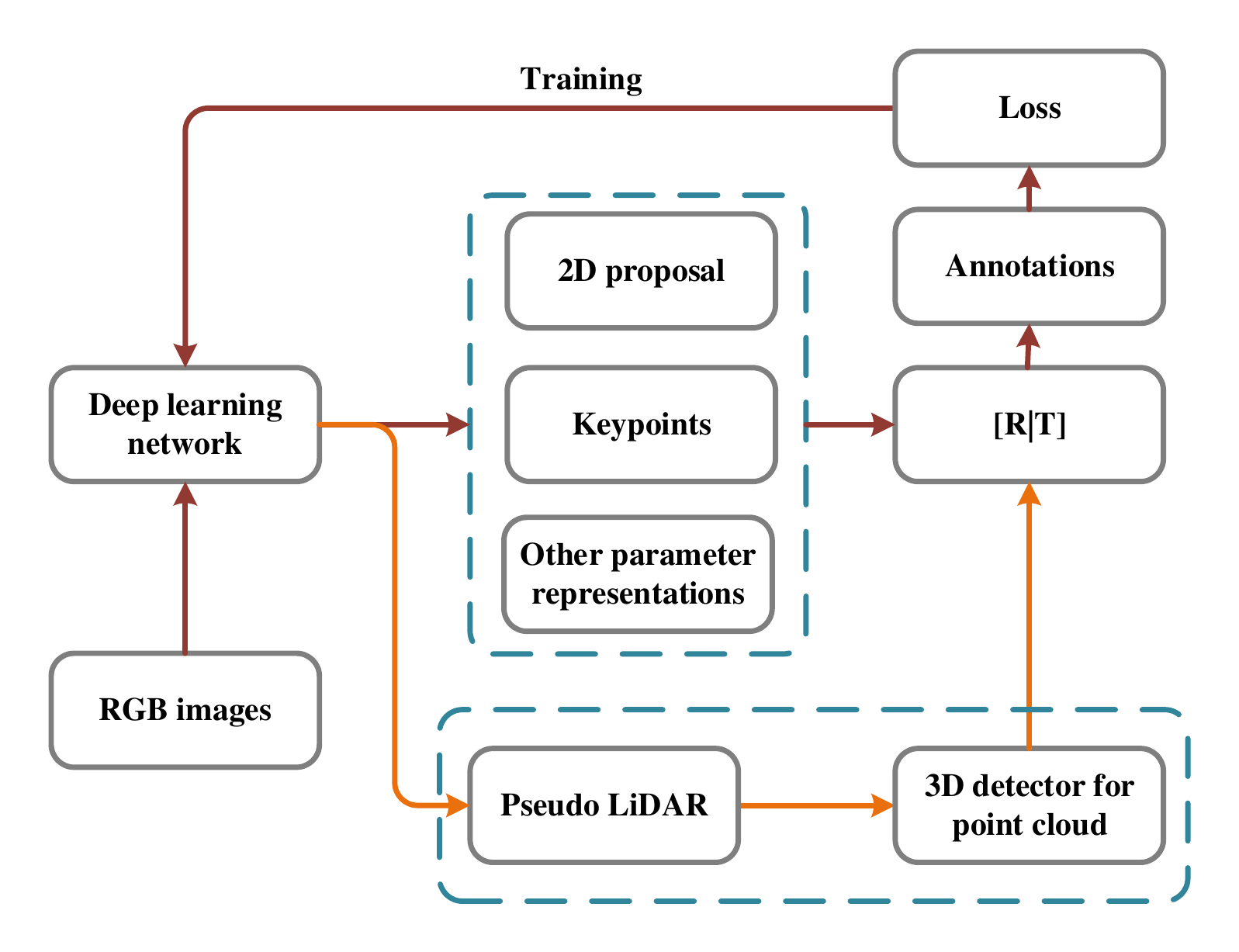}\label{category3ddetection}} 
	\subfigure[6D pose detection methods]{\includegraphics[width=2in]{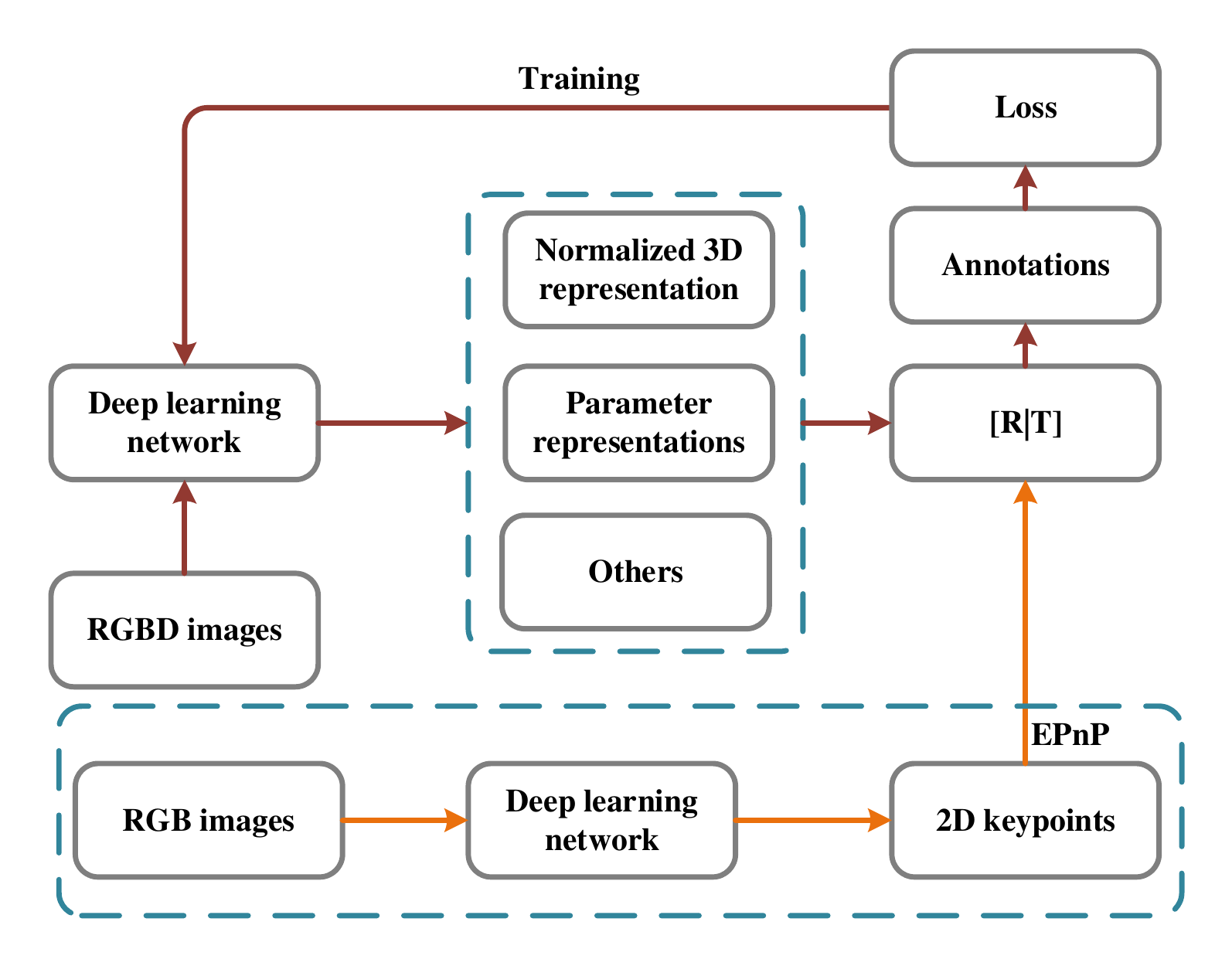}\label{category6dpose}} 
	\vspace{-0.2in} 
	\caption{ Overall schematic representation of category level monocular detection methods} 
	\vspace{-0.1in} 
\end{figure*} 

\subsection{Category-Level Monocular 3D Object Detection}

\begin{table*}
	\centering
	\resizebox{\textwidth}{40mm}{
		\begin{tabular}{r |c|c|c|c|c|c }  
			\hline 	
			\multirow{2}*{Methods}&\multirow{2}*{Years}&\multirow{2}*{Types} &\multirow{2}*{Depth Estimation} & [val/val2/test]&[val/val2/test]&[val/val2/test]\\
			
			& &  & & Easy&Moderate&Hard\\
			\hline 	
			Mono3D \cite{chen2016monocular}&2016&Others &No&2.53/ - / - &2.31/ - / - & 2.31/ - / -\\
			
			Deep3Dbox \cite{mousavian20173d}&2017&2D proposal&No&5.85/ - / - &4.10/ - / - &3.84/ - / - \\
			
			GS3D \cite{li2019gs3d}&2019&2D proposal&No&13.46/11.63/4.47 &10.97/10.51/2.90 &10.38/10.51/2.47\\
			
			FQNet \cite{liu2019deep}&2019&2D proposal &No&5.98/5.45/2.77&5.50/5.11/1.51&4.75/4.45/1.01\\

			ROI-10D \cite{manhardt2019roi}&2019&2D proposal&No&10.12/ - / - &1.76/ - /&-1.30/ - / - \\
			
			MonoGRNet \cite{qin2019monogrnet}&2019&2D proposal&No&13.88/ - /9.61&10.19/ - /5.74&7.62/ - /4.25\\
			
			MonoDIS \cite{simonelli2019disentangling}&2019&2D proposal&No&18.05/ - /10.37&14.96/ - /7.96&13.42/ -/6.40\\

			M3D-PRN \cite{brazil2019m3d}&2019&2D proposal&No&20.27/20.40/14.76&17.06/16.48/9.71&15.21/13.34/7.42\\
			
			SMOKE \cite{liu2020smoke}&2019&Keypoints&No& - /14.76/14.03& - /12.85/9.76& - /11.50/15.28\\
			
			Jorgensen et al. \cite{jorgensen2019monocular}&2019&Keypoints&No& 14.52/9.45/11.74&13.15/8.42/9.58&11.85/7.34/7.77\\
			
			Pseudo-LiDAR \cite{wang2019pseudo}
			&2019&Psudeo-LiDAR&Yes&28.2 / - / - &18.5/ - / - & 16.4/ - / -\\

			Mono3D-PliDAR\cite{weng2019monocular}
			&2019&Psudeo-LiDAR&Yes&31.5  / - / 10.76 &21.0/ - / 7.50 & 17.5/ - / 6.10\\

			AM3D\cite{ma2019accurate}
			&2019&Psudeo-LiDAR&Yes&32.23  / 28.31 / 16.50 &21.09/ 15.76 / 10.74 & 17.26/ 12.24 / 9.52\\
			
			DA-3Ddet\cite{fengmonocular}
			&2020&Psudeo-LiDAR&Yes&33.4 / - / 16.8 &24.0/ - / 11.5 & 19.9/ - / 8.9\\
			
			PatchNet\cite{ma2020rethinking}
			&2020&Psudeo-LiDAR&Yes&35.1  / 31.6 / 15.68 &22.0/ 16.8 / 11.12 & 19.6/ 13.8 / 10.17\\
			
			D4LCN\cite{ding2020learning}
			&2020&Psudeo-LiDAR&Yes&26.97 / 24.29 / 21.71 &24.0/ 19.54 / 11.72 & 18.22/ 16.38 / 9.51\\

			FADNet\cite{gao2020monocular}
			&2020&Keypoints&No& 23.98/ - /16.37&16.72/ - /9.92&16.72/ - /8.05\\

			MoNet3D\cite{zhou2020monet3d} 
			&2020&Keypoints&No& 22.73/ - / - & 16.73/ - / - &15.55/ - / - \\
			
			RTM3D \cite{li2020rtm3d} 
			&2020&Keypoints&No& 20.77/ - / 14.76 & 16.86/ - / 9.71 &16.63/ - / 7.42 \\

			UR3D \cite{shi2020distance} &2020&Others&Yes& 28.05/26.30/15.58& 18.76/16.75/8.61&16.55/13.60/6.00\\
			RAR-Net  \cite{liu2020reinforced} &2020&Others&No& 23.12 / - /16.37& 19.82 / - /11.01&16.19 / - /9.52\\
			
			kinematic3D \cite{brazil2020kinematic}  &2020&Others(video)&No& - / - /19.07& - / - /12.72&- / - /9.17\\

			Li et al. \cite{li2021monocular} &2021&Keypoints&No& 22.50/22.71/16.73 &19.60/17.71/11.45 &17.12/16.15/9.92\\
			
			Liu et al. \cite{liu2021ground}&2021&Others&Yes& 23.63/ - /16.65 & 16.16/ - /13.25 &12.06/ - /9.91\\
			
			CaDDN\cite{reading2021categorical}
			&2021&Psudeo-LiDAR&Yes&23.57 / - / 19.17 & 16.31/ - / 13.41 & 13.84/ - / 11.46\\
			
			OCM3D \cite{peng2021ocm3d}&2021 &Psudeo-LiDAR&Yes&23.65 / - / 17.48 & 17.75/ - / 10.44 & 15.93/ - / 7.87\\
			
			Peng et al. \cite{peng2021lidar}&2021 &Psudeo-LiDAR&Yes&26.17 / - / 22.73 & 19.61/ - / 14.82 & 15.93/ - / 12.88\\
				FOCS3D \cite{wang2021fcos3d}&2021&Keypoints &Yes& 13.9 / - / - & 11.6 / - / - & 11.0 / - / -\\
			PGD \cite{wang2021probabilistic}&2021&Pseudo-Lidar &Yes& 24.4 / - / - & 18.3 / - / - & 16.9 / - / -\\
			DD3D \cite{park2021pseudo}&2021 &Psudeo-LiDAR&Yes& - / - / 23.22 & - / - / 16.34 & - / - / 14.20\\
			\hline
			
	\end{tabular}}
	
	\caption{Perfromance of  category level monocular 3D object detection methods on KITTI3D dataset.}
	\vspace{-0.3in} 
	\label{kitti}
\end{table*}

\begin{table*}
	\centering
		\begin{tabular}{r |c|c|c|c|c}  
			\hline 	
			Methods&Years&Types &Depth Estimation &mAP&NDS\\
			
			\hline 	
			Mono3D \cite{chen2016monocular}&2016&Others &No&36.6 &0.43\\
			CenterNet \cite{duan2019centernet} &2019&Keypoints &No&33.8 &0.40\\
			MonoDIS \cite{simonelli2019disentangling}&2019&2D proposal&No&30.4&0.38\\
			FCOS3D \cite{wang2021fcos3d} &2021&Keypoints &No&35.8&0.43 \\
			PGD \cite{wang2021probabilistic}&2021&Pseudo-Lidar &Yes&38.6&0.45 \\
			DD3D \cite{park2021pseudo}&2021&Pseudo-Lidar &Yes&41.8&0.48 \\

			\hline		
	\end{tabular}
	\caption{Perfromance of  category level monocular 3D object detection methods on Nuscenes dataset.}
	\vspace{-0.4in} 
	\label{nuscenes}
\end{table*}

Category-level monocular 3D object detection needs to predict 7 degrees of freedom (7Dof) pose configurations, including rotation $\mathcal{R} \in SO(1)$ (i.e. only yaw needs to be predicted), translation $\mathcal{T} \in R^3$ and object size $\mathcal{S} \in R^3$ . No CAD models are available during training and testing.  Category-level monocular 3D object detection is of great importance to autonomous driving scenarios. It is more concerned about the accuracy of translation prediction, whereas the accuracy of rotation prediction can be relaxed accordingly.  Point clouds collected by LiDAR and monocular RGB images are the most commonly used data formats. The former is out of the scope of this study because this work focuses on monocular object pose detection, and readers are referred to \cite{guo2020deep,fernandes2021point,arnold2019survey} for a detailed introduction. Figure \ref{category3ddetection} shows an overall schematic representation of category-level monocular 3D object detection methods that take RGB images as input. 

To the best of our knowledge, Mono3D \cite{chen2016monocular} is the first deep learning method that achieves this goal. It focuses on generating 3D object proposals using instance segmentation, object contour, and ground-plane assumption. However, the ground-plane assumption does not always hold in several scenarios, and utilizing many auxiliary supervision signals to train a model would burden the model during training. 

\textbf{2D proposal-based methods}. Soon, Deep3Dbox \cite{mousavian20173d} is proposed as an external state-of-the-art object detector to generate 2D proposals and process the cropped proposals within a deep neural network to estimate 3D dimensions and orientation. In Deep3Dbox, the relationship between the predicted 2D boxes and the projected 3D boxes on the image plane is exploited (by aligning projections of 3D box corners with lines of corresponding 2D boxes) in post optimization to help calculate 3D parameters. The 3D box cannot be predicted in an end-to-end manner using the deep learning network, which is a certain limitation. Next, GS3D \cite{li2019gs3d} leverages the off-the-shelf 2D object detector Faster-RCNN \cite{ren2016faster}  to  obtain a coarse cuboid using each predicted 2D box as guidance, hence achieving end-to-end training. 

Though an end-to-end framework, GS3D highly relies on the assumption that the top center of the object 3D box has a stable projection on the 2D plane (i.e., it is very close to the top midpoint of the 2D box), and so does the bottom center of the 3D box. However, the assumption is not strictly mathematically reasonable. M3D-PRN \cite{brazil2019m3d} is proposed to use the depth-aware convolution to generate 2D and 3D object proposals simultaneously and use 2D-3D geometric constraints as post processing to improve precision. In M3D-RPN, several geometric constraints, such as corresponding 2D boxes of the 3D anchors with the same size that should satisfy the law of near-large and far-small, are introduced to bound the 2D and 3D anchors for better prediction results. However, predefining 2D-3D bounded anchors is a difficult problem because exhausting all possibilities is difficult. FQNet \cite{liu2019deep} is proposed to infer 3D IoU between the predicted 3D box and ground-truth box to predict the most likely candidate 3D bounding box. Therefore, FQNet can learn and reason about 3D spatial relationships from 2D projections. However, FQNet requires image patches to be cropped and resized, which is inefficient, especially when an image has many objects. By contrast, ROI-10D \cite{manhardt2019roi} is more straightforward because it directly utilizes features of region-of-interest (RoI) to lift 2D detection to 3D with no image patch cropping required. The above-introduced methods predict 3D boxes by leveraging their relationship with the corresponding 2D bounding boxes.

MonoGRNet \cite{qin2019monogrnet} directly optimizes the 3D box in an end-to-end manner and achieves accurate 3D object detection by fusion of 2D detection, instance depth estimation, 3D location estimation, and local corner regression. The depth estimation in MonoGRNet refers to only predicting the depth of centers of 3D bounding boxes. Therefore, no additional ground-truth depth maps are needed as labels to train the network, avoiding the limitations of the additional annotation cost. Moreover, MonoDIS \cite{simonelli2019disentangling} discovers that isolating the group parameters during training would benefit the network. Inspired by this observation, MonoDIS is trained by disentangling the 3D parameters into rotation, translation, and object size, and isolated loss functions are used to supervise learning. More importantly, the disentangling training mode can be integrated into any monocular 3D object detection model.

The 2D proposal-based methods are inspired by 2D object detection. They are easy to understand and could achieve excellent performance. However, their main limitation is they incorporate a 2D proposal network as the base architecture, which causes the bottlenecks of inference time. 

\textbf{Keypoints-based methods}. To overcome the above problem, instead, SMOKE \cite{liu2020smoke} is proposed to predict the 3D keypoints of object 3D bounding boxes directly through a concise single-stage framework. Then, FADNet \cite{gao2020monocular} improves SMOKE by using several ConvGRU layers \cite{siam2017convolutional} to learn parameters related to calculate 3D keypoints. Moreover, it suggests learning a depth hint vector to better perceive the depth difference between different pixel rows better, which coincides with the idea of depth-convolution in M3D-RPN. However, the large network structure of FADNet makes achieving real-time performance difficult. Similarly, FCOS3D \cite{wang2021fcos3d} adopts a center-based paradigm to decouple the 3D object detection task as a multitask learning problem in the detection head. In this work, objects are distributed to different feature levels considering their 2D scales and assigned only according to the projected 3D-center for the training procedure. MoNet3D \cite{zhou2020monet3d} is also a keypoints-based method. It is highly dependent on the assumption that if two objects are horizontally close in the image and have the same depth, they locate close in 3D space. 

Instead of directly predicting 3D keypoints, RTM3D \cite{li2020rtm3d} first predicts 2D keypoints and then uses geometric constraints to lift and refine 2D keypoints to 3D box-related parameters. Compared with previous methods, RTM3D has achieved a great improvement in accuracy and speed. However, the post lifting and refinement process is nondifferentiable, making implementing end-to-end prediction impossible. Recently, Li et al. \cite{li2021monocular} successfully solved the problem by designing a differentiable refinement module that can precisely predict 3D boxes. Similar to RTM3D, Jorgensen et al. \cite{jorgensen2019monocular} decompose the 3D box and the 2D box into 26 predictors. After predicting these 26 parameters, the rough 3D bounding box is obtained. Then a least square algorithm is used to refine the 3D bounding box, with 3D IoU as the best optimization target. However, the least square algorithm they use is also nondifferentiable.

The advantages of keypoints-based methods are fast and accurate. However, the final detection accuracy is highly dependent on the keypoint detection accuracy, that is, the detection is not achieved end-to-end. Owing to inaccurate keypoint detection, these methods are easier to attack by occlusion and truncation.

\textbf{Pseudo-LiDAR methods}. Unlike previous RGB image-based methods, the concept of pseudo-LiDAR is proposed by PL-MONO (Pseudo-LiDAR) \cite{wang2019pseudo} that introduces depth information into 3D object detection. Instead of directly estimating 3D bounding boxes from the scene, pseudo-LiDAR predicts the depth $Z(u,v)$ of each image pixel $(u,v)$. The resulting depth map $Z$ is then projected into a 3D point cloud, and a pixel $(u,v)$ is transformed to $(x,y,z)$ by
$$z=Z(u,v), x=\frac{(u-c_u)\times z}{f_U}, y=\frac{(v-c_V)\times z}{f_V}$$
where $(c_U,c_V)$ represents the camera center; $f_U$ and $f_V$ denote the horizontal and vertical focal lengths, respectively. The pseudo-LiDAR points can then be regarded as LiDAR signals, and any off-the-shelf LiDAR-based 3D object detector can be applied to them. By making use of state-of-the-art depth estimation algorithms, the simple yet effective PL-MONO (Pseudo-LiDAR) method has achieved promising improvement on the monocular 3D object detection task benchmarked on KITTI3D dataset.

Although PL-MONO (Pseudo-LiDAR) has greatly promoted performance,a substantial performance gap exists between using pseudo-LiDAR and using real LiDAR. Afterward, Weng and  Kitan \cite{weng2019monocular} propose model Mono3D-PLiDAR and point out that the noise in pseudo-LiDAR data severely limits the detection performance of the current method. Guided by the observed misleading and long-tail issues in the noisy pseudo-LiDAR, a 2D-3D bounding box consistency loss is proposed as an additional supervision signal in Mono3D-PLiDAR to solve the problem. Pseudo-LiDAR++ \cite{you2019pseudo} reckons that the quality of depth estimation plays an essential part in pseudo-LiDAR related methods, especially for the depth accuracy of far objects.  Therefore, a stereo network architecture, as well as corresponding loss functions, is designed to estimate the depth of faraway objects.  Next, E2E-PLiDAR \cite{qian2020end} is proposed as an end-to-end pipeline-based on differentiable Change of Representation modules that allow back-propagation throughout all layers to preserve the modularity and compatibility of pseudo-LiDAR. 

The above methods all rely on off-the-shelf 3D object detectors for predicting 3D bounding boxes. Ma et al. \cite{ma2020rethinking} analyze the effect of depth data representation on performances and discover that performance improvement comes from 3D coordinate representation rather than 3D detectors. They propose PatchNet, which improves performance over previous pseudo-LiDAR models by integrating the 3D coordinates as additional input channels rather than treating them as point clouds. PatchNet attempts to improve detection performance from the perspective of obtaining better representations, whereas authors of DA-3Ddet \cite{fengmonocular} attempt to solve the problem of learning more precise features. In DA-3Ddet,  the feature from the image-based pseudo-LiDAR domain is adapted to the real LiDAR domain, contributing to a novel domain adaptation-based monocular 3D object detection framework.  A Context-Aware Foreground Segmentation module is also introduced in DA-3Ddet to solve the inconsistency between the foreground mask of pseudo-LiDAR and real LiDAR caused by inaccurately estimated depth. 

In contrast to the methods mentioned above that focus on improving depth feature quality or representation quality,  authors of AM3D \cite{ma2019accurate} claim that RBG features from the original image should not be ignored either. Therefore, they improve the point cloud representation by adopting a multimodal feature fusion module to integrate complementary RGB features into the pseudo-LiDAR pipeline. Reading et al. \cite{reading2021categorical} also research how to combine image features in model CaDDN, where bird’s-eye-view (BEV) images are used. Specifically, they use each pixel's predicted categorical depth distribution to project rich contextual feature information to appropriate depth intervals in 3D space. Then BEV projection of the 3D presentation is fed into a single-stage detector to produce the final detection result. 

The above methods all focus on learning better features, while Ding et al. \cite{ding2020learning} investigate how to learn better convolutional kernels in D4LCN. They propose a new local convolutional network (LCN). Instead of learning global kernels to apply to all images, the convolutional kernels are automatically learned from image-based depth maps and locally applied to each pixel and channel of individual image sample, narrowing the gap between image and 3D point cloud. Recently, Simonelli\cite{simonelli2020demystifying} observes that the validation results published by existing pseudo-LiDAR-based methods are substantially biased. As a result, the mechanism for predicting a 3D confidence score is further introduced to address the problem. Then, authors of OCM3D \cite{peng2021ocm3d} again find that current methods cannot handle the noisy nature of the monocular pseudo-LiDAR point cloud.  Therefore, they improve the performance by building voxels for each object proposal, allowing the noisy point cloud to be organized effectively within a voxel grid.  PGD \cite{wang2021probabilistic} believes that the accuracy of depth estimation is more important. Therefore, it incorporates a probabilistic representation to capture the uncertainty in depth estimation.

Nearly at the same time, Peng et al. \cite{peng2021lidar} argue that implicit utilizing LiDAR point clouds to train depth estimation models cannot fully explore their capabilities and consequently lead to suboptimal performances. To tackle this, they directly use real LiDAR point clouds to guide the training of monocular 3D detectors rather than use repredicted depth,  which substantially improves performance. In Peng et al. \cite{peng2021lidar}, real LiDAR point clouds are only required during training, which is the same as training a depth estimator.  Similarly, DD3D \cite{park2021pseudo} claims that predicting depth as an intermediate process is unnecessary. Therefore, it proposes to pretrain the model on depth estimation and then finetune the model on monocular 3D object detection. Unexpectedly, this also achieves state-of-the-art performance, demonstrating that how to utilize depth annotation better for training 3D object detection is worth researching. 

Briefly, pseudo-LiDAR methods are advanced and much more accurate than other methods benefiting from the additional estimated depth information. However, most of the existing methods are much more computationally expensive than others because they need to estimate depth, transform depth into point clouds and detect objects hidden in point clouds, for a total of three steps.


\textbf{Other methods}. Beyond the above methods, Shi et al.  \cite{shi2020distance} propose a distance-normalized unified representation (DNRU) in model UR3D to help the model learn a unified representation for objects at different depths. However, their method requires estimating pixel depth to construct the DNRU simultaneously, and most of the detection accuracy gains come from the estimated depth, which is neither effective nor efficient. Liu et al. \cite{liu2021ground} introduce a method that can leverage the ground information as priors. It removes nonground anchors to reduce redundancy and votes the ground pixel below the current pixel to learn geometric cues. Although intuitive, voting for ground pixels is not easy, especially when interference factors such as occlusion exist. It also needs to predict depth map for improving performance.

In addition to the monocular 3D detection method that is defined as taking a single image as input, Brazil et al. \cite{brazil2020kinematic} propose model kinematic3D, which leverages 3D kinematics from monocular videos to improve the overall localization precision of the monocular 3D object detection task. It also produces useful scene dynamic byproducts (i.e., ego-motion and per-object velocity) that benefit autonomous driving. Beyond methods focusing on designing end-to-end network architectures,  a post optimization that can be integrated into any monocular detectors is proposed in RAR-Net \cite{liu2020reinforced}. This method starts with an initial 3D box prediction and then gradually refines it toward the ground truth, with only one 3D parameter changed in each step. The progress is optimized by reinforcement learning. The limitation is RAR-Net requires optimization for multiple steps, which increases additional computational cost.

Till now, the recent state-of-the-art monocular 3D object detection methods have been introduced. Table \ref{kitti} presents an overview and a complete comparison of representative works. The results are about the performance of these methods on KITTI val1, val2, and test split. The mAP@IoU75 of \emph{Car} class, which is the most widely concerned category in monocular 3D object detection, is reported. The results about val1 split and val2 split are reported using the $mAP_{11}@IoU70$ metric to provide a complete comparison for all published related works. The results about the test split are reported using the $mAP_{40}@IoU70$ metric, which is the latest standard evaluation metric of KITTI's official leaderboard.  Table \ref{nuscenes} shows a comparison between current state-of-the-art methods on Nuscenes datasets. The metrics are mAP and NDS.

\subsection{Category-Level Monocular 6D Pose Detection}

\begin{table*}
	\resizebox{\textwidth}{25mm}{
		\centering
		\begin{tabular}{c |r|c|c|c|c|c|c|c }  
			\hline 	
			Data& Methods &Input&To pose& mAP@IoU50  & mAP@IoU75&  mAP@5\degree 5cm& mAP@10\degree 2cm& mAP@10\degree 5cm\\
			\hline 	
			\multirow{6}*{CAMERA25} &NOCS(2019)\cite{wang2019normalized} &RGBD	&Aligning& 83.9 & 69.5& 40.9& 48.2& 64.6  \\
			&SPD(2020)\cite{tian2020shape} 	&RGBD&Aligning&92.2 & 83.1& 59.0& 73.3& 81.5     \\
			&CPS(2020)\cite{manhardt2020cps++} &RGBD&Regressing	& 70.4 & - & 33.6& -& - \\
			&CPS+ICP(2020)\cite{manhardt2020cps++}&RGBD&Aligning& 63.4 & - & 42.8& -& 63.8 \\
			&DualPoseNet(2021)\cite{lin2021dualposenet}&RGBD&Regressing&92.4 &86.4& 70.7& 77.2&84.7 \\
			&ACR-Pose(2021)\cite{fan2021acr}&RGBD&Aligning&93.8 &89.9& 74.1& 82.6&87.8 \\
			&SGPA(2021)\cite{chen2021sgpa}&RGBD&Aligning&93.2 &88.1& 74.5& 82.7&88.4 \\			
		&Lee et al(2021)\cite{lee2021category}&RGB&Aligning&32.4 &5.1& - & - &-  \\

			\hline 	
			\multirow{6}*{REAL275} &NOCS(2019)\cite{wang2019normalized}&RGBD&Aligning& 78.0 & 30.1& 10.0& 13.8& 25.2    \\
			&SPD(2020)\cite{tian2020shape} 	&RGBD&Aligning& 77.3 & 53.2& 21.4& 43.2& 54.1     \\
			&CASS(2020)\cite{chen2020learning}&RGBD &Regressing	& 84.2 & 77.7& 23.5& 58.0& 58.3 \\
			&CPS(2020)\cite{manhardt2020cps++}&RGBD&Regressing 	& 72.6 &  - & 25.8& -& - \\
			&CPS+ICP((2020))\cite{manhardt2020cps++}&RGBD&Rligning 	&  72.8 & - & 25.2& -& - \\
						&FS-Net(2021)\cite{chen2021fs}&RGBD&Regressing&92.2 &63.5& 28.2& -&60.8 \\
			&DualPoseNet(2021)\cite{lin2021dualposenet}&RGBD&Regressing&79.8 &62.2& 35.9& 50.0&66.8 \\

		&ACR-Pose(2021)\cite{fan2021acr}&RGBD&Aligning&82.8 &66.0& 36.9& 54.8&65.9 \\	
		&SGPA(2021)\cite{chen2021sgpa}&RGBD&Aligning&80.1 &61.9& 39.6& 61.3&70.0 \\
		&Lee et al(2021)\cite{lee2021category}&RGB&Aligning&23.4 &3.0& - & - &-  \\
			\hline 	
	\end{tabular}}
	\caption{Performance of  category level  monocular 6D pose detection methods on NOCS dataset.}
	\label{nocs}
	\vspace{-0.3in} 
\end{table*}

\begin{table*}
	\centering
	\resizebox{\textwidth}{5mm}{
		\begin{tabular}{r |c|c|c|c|c|c|c|c|c|c|c}  
			\hline 	
			Methods& Input & How to pose& bike	&book&	bottle&camera&	cereal\_box& chair&	cup&	 laptop	&shoe\\
			\hline
			MobilePose(2020)\cite{hou2020mobilepose}&RGB&EPnP&0.3486&	0.1818&	0.5449&	0.4762&	0.5496&	0.7112&	0.3722	&0.5548&	0.423\\
			MobilePose v2(2020)\cite{ahmadyan2020objectron}&RGB&EPnP&0.6127	&0.5218&	0.5744&	0.8016&	0.6272&	0.8505&	0.5388&	0.6735&	0.6606\\
			\hline 	
			Lin et al. \cite{lin2021single}&RGB&EPnP&0.6419& 0.5565& 0.8021& 0.7188 &0.8211& 0.8471 &0.7704& 0.6766& 0.6618\\
			\hline 	
			
	\end{tabular}}
	\caption{Performance of  category level monocular 6D pose detection methods on Objectron dataset.}
	\label{objectron}
	\vspace{-0.35in} 
\end{table*}
In category-level monocular 6D pose detection task, 9 degrees of freedom (9Dof) pose configurations including rotation $\mathcal{R} \in SO(3)$, translation $\mathcal{T} \in R^3$, and object size $\mathcal{S} \in R^3$ need to be predicted. No CAD model is available to help ease the estimation task. Strictly speaking, it should be named as 9Dof object pose detection. To conform to customary convention, it is still called category-level monocular 6D pose detection in this survey. The main challenge of this task is how to endow the model the ability to deal with intraclass variation \cite{sahin2019instance}. Compared with the category-level monocular 3D object detection task, this task requires the model to predict another 2 degrees of freedom w.r.t rotation. Specifically, the the pitch and roll angles except for the yaw angles need to be predicted. Intuitively, the difference seems not much, but learning the entire SO(3) space is much more difficult than just learning the SO(1) space, especially when only RGB images can be used as input.

Sahin et al. \cite{sahin2018category} are the first to propose the concept of category-level pose estimation. In their work, a part-based Random Forest is introduced, where a category of CAD instances is decomposed into parts and represented with skeletons to train the random forest. During inference, parts of an instance are organized in the form of several trees to hypothesize 6Dof poses. Their methods highly rely on the geometric characters of a category/instance. Therefore, it mainly deals with depth input. Moreover, the dataset they mainly focus on is quite simple and is limited to an ideal environment (like instance images with no background).  

\textbf{Detecting by aligning}. Wang et al. \cite{wang2019normalized} introduce a shared canonical representation for all possible object instances within a category named normalized object coordinate space (NOCS), which, strictly speaking, is the first deep learning method for category-level monocular 6D pose detection. In Wang et al. \cite{wang2019normalized}, the RGBD patch of the target object is first segmented using the detection results of an off-the-shelf 2D object detection/segmentation method Mask-RCNN \cite{he2017mask}. Then the patch is learned to predict the NOCS representation using a region-based neural network. Next, the 6Dof pose and object size can be recovered using the Umeyama algorithm \cite{umeyama1991least} by aligning the predicted NOCS in the camera coordinate system to the canonical NOCS in the object coordinate system. Another contribution of Wang et al.\cite{wang2019normalized} is that they have released the  NOCS dataset (including CEMERA25 and REAL275), which is now the most widely used benchmark for category-level monocular 6D pose detection.  

Wang et al.'s work \cite{wang2019normalized}  performs well. However, inferring the object pose by only predicting the NOCS presentation is not an easy task because intra class variation is very large. Therefore, SPD \cite{tian2020shape} is proposed to handle the intra-class variations by learning a deformation field from the pre-learned shape prior. SPD infers the dense correspondences between the depth observation of the object instance and the reconstructed NOCS representation to estimate the 6Dof object pose and size jointly. The following work ACR-Pose \cite{fan2021acr} states that reconstructing the canonical NOCS representation is critical. Therefore, an adversarial training scheme is proposed in this work to help the network reconstruct high-quality canonical representations. SGPA \cite{chen2021sgpa} proposes a methods that utilizes Transformers to fuse RGB features, depth features , and shape prior features better for the canonical NOCS representation reconstruction. Though all of them have improved the performance, they still need the Umeyama \cite{umeyama1991least} post optimization algorithm and can not predict the poses of objects in an end-to-end manner. Moreover, generalizing to RGB-only scenarios is difficult for them.  To generalize category-level methods to RGB-only scenarios, Lee et al. \cite{lee2021category} propose to predict the depth and the NOCS representation simultaneously. The depth is recovered by first predicting the object mesh and then rendering the mesh into a depth map. Though RGB-only detection is achieved, the performance of \cite{lee2021category} is far behind RGBD-based the above counterparts.

In conclusion, detecting by aligning benefits from constructing a predefined canonical representation increases pose detection accuracy. However, they are often slow, and achieving real-time performance is difficult for them due to complex reconstruction.

\textbf{Detecting by regressing}. Appropriately, Chen et al. \cite{chen2020learning} propose CASS, where a canonical shape space that models the latent space of canonical 3D shapes with a normalized pose is learned to tackle intra class shape variations. In their model, shape-dependent features and pose-dependent features are contrasted and fused to regress the object pose and size directly. CPS \cite{manhardt2020cps++} is another model that directly regresses 6Dof pose and size, which predicts 6Dof pose and size using RoI features after RoI Align \cite{he2017mask}. In CPS, a shape encoding is learned to recover the point cloud using a PointNet \cite{qi2017pointnet} decoder. Thus, the features learned by the encoder network can be improved with additional supervision during training. However, the performance of CPS is poor if it directly predicts the pose and size without using ICP as a refinement.

More recently, DualPoseNet \cite{lin2021dualposenet} is proposed to improve the performance of detecting by regressing methods further. The improvement of DualPoseNet comes from the used spherical convolution \cite{esteves2018learning} and its dual explicit-implicit pose decoder architecture. Spherical convolution is used to learn SO(3) equivalent representations to recover the rotation easier using features learned from it. The explicit decoder and the implicit decoder are used to directly predict pose/size directly and reconstruct canonical point cloud representation of the target object, respectively. Then, a refinement algorithm is proposed in DualPoseNet by considering the consistency between the explicit decoder and the implicit decoder.  

In addition to DualPoseNet, Chen et al. \cite{chen2021fs} recently propose FS-Net to predict object pose and size directly. In FS-Net, an orientation-aware autoencoder with 3D graph convolution is used for latent feature extraction. The 3D graph convolution is scale and translation invariant. Therefore, the learned features are only aware of rotation, resulting in a more precise rotation recovery. It recovers the scale and size by learning residuals between the scale and size of the reconstructed point cloud and the ground truth, reducing research space and easing the task. Chen et al. \cite{chen2021fs} propose a 3D Deformation Mechanism in their work. They generate new training examples by enlarging, shrinking, or changing the area of several surfaces of the box-cages to deform rigid objects, which increases the abundance of intra class variations and prompts the model to learn more discriminative features. Detecting by regressing methods demonstrates the potential of extension to RGB-only scenarios. However, minimal effort has been paid towards this direction. 

Although  detecting by regressing methods overcomes the limitation of reconstructing first  before pose estimation, their performance is relatively poor because they currently do not use any prior information in their deep networks.

\textbf{Detecting by keypoints}. The above models consist of the recently state-of-the-art category-level monocular 6D pose detection methods. However, they share a common drawback, that is, they all mainly focus on utilizing RGBD data as input that limits the applicated scenarios. Specifically, the above methods cannot apply to the AR-related applications on mobile phones because obtaining depth information through portable devices is difficult. Moreover, complex network architecture and time-consuming post optimization make achieving real-time performance difficult. Therefore, MobilePose \cite{hou2020mobilepose} is proposed. MobilePose is built upon the lightweight MobileNet v2 \cite{sandler2018mobilenetv2}. It predicts the 2D keypoints of the object in the image. It only takes RGB images as input. Then, the up-to-scale 9Dof object bounding box is recovered by solving the EPnP \cite{lepetit2009epnp} problem.

Lately, Ahmadyan et al. \cite{ahmadyan2020objectron} release the Objectron dataset they used and introduce MobilePose v2, a two-stage category-level 6D pose detection pipeline built upon MobilePose.  In MobilePose v2, SSD \cite{liu2016ssd} is first used to detect 2D object patches in the image. Then, detected patches are cropped, resized, and input into an EfficientNet-like \cite{tan2019efficientnet} network to predict 2D keypoints. Then, the same EPnP algorithm is used as in MobilePose to recover the up-to-scale 9Dof object bounding box. MobilePose and MobilePose v2 can achieve real-time performance and can be applied to terminals such as mobile phones.  Later, Lin et al. \cite{lin2021single} propose a novel single-stage method that outperforms the two-stage  MobilePose v2. Recently, in Lin et al. \cite{lin2021single}, the object center and keypoints are predicted by different heat maps and the relative cuboid dimensions directly predicted by the network. Then, they use a PnP algorithm to solve the pose and object size. However, their result is still an up-to-scale solution.

Methods detecting by keypoints are lightweight and easy to be deployed. However, the final solutions are up-to-scale (i.e., they cannot estimate the absolute depth of the object from the camera), which may be the main drawback of this kind of method, because they mainly focus on serving AR.

\textbf{Other methods}. In addition to the above discussed major kinds of methods, several methods focus on designing more general solutions or handling more challenging cases. For example, Chen et al. \cite{chen2020category} propose to use an analysis-by-synthesis strategy for category level 6D pose detection. Specifically, they propose a gradient-based fitting procedure with a parametric neural image synthesis module to implicitly represent entire object categories' appearance, shape, and pose implicitly. Moreover, rendering can be achieved without CAD models. Pose can be predicted by comparing the decoded image and the observed real image. Though this method achieves the goal of only taking RGB data as input, the performance drops substantially when depth information is not available. Moreover, it may be sensitive to initialization and time consuming due to refining for multiple iterations. 

In contrast to predicting the pose of rigid objects, Li et al. \cite{li2020category} propose Articulation-aware Normalized Coordinate Space Hierarchy (ANCSH), a canonical representation for different articulated objects in a given category for detecting object pose of articulated objects.  The task involved in this work requires the model to predict per part 6D poses, 3D scales, joint parameters (i.e., type, position, axis orientation), and joint states (i.e., joint angle). Although ANCSH has achieved good results only by using depth information, how to use the more feature-rich RGB information has not been studied in this work.

Till now, the recent state-of-the-art category-level monocular 6D pose detection methods have been introduced.  The above methods are mostly benchmarked on NOCS dataset (CEMERA25 and REAL275) with the evaluation metrics  mAP@IoU25, mAP@IoU50, mAP@5\degree 5cm, mAP@10\degree 2cm, mAP@10\degree5cm to measure performances jointly. Table \ref{nocs} reports their main results on RGBD data. Table \ref{objectron} presents the results of MobilePose and MobilePose v2 on Objectron dataset. The evaluation metric is mAP@IoU50.  

\begin{figure}
	\centering  
	\includegraphics[width=0.99\textwidth]{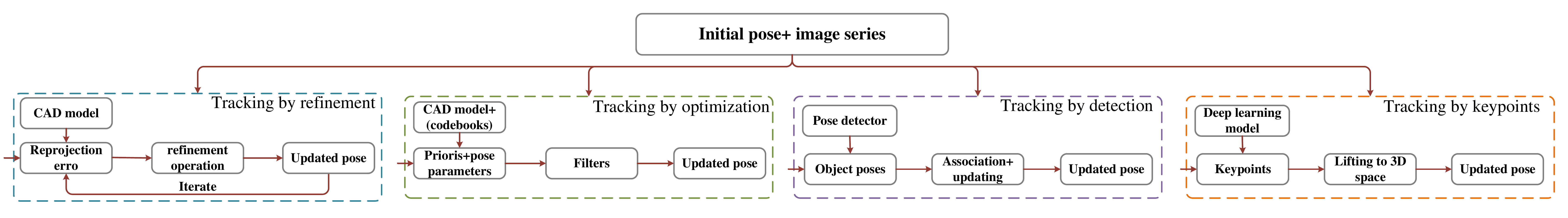}  
	\vspace{-0.2in}
	\caption{Overall schematic representation of  monocular object pose tracking.}
	\label{tracking}
	\vspace{-0.15in}
\end{figure}

\section{Monocular Object Pose Tracking}
\label{sectracking}
In this section, monocular object pose tracking methods are introduced. According to whether CAD models are available, related methods are classified into \textbf{Instance-Level Monocular Object Pose Tracking} and \textbf{Category-Level Monocular Object Pose Tracking}. Figure \ref{tracking} shows an overall schematic representation of  monocular object pose tracking. 

\subsection{Instance-Level Monocular Object Pose Tracking}
Instance level monocular object pose tracking requires us to predict 6 degrees of freedom(6Dof) pose of a sequence of images (i.e., a video). The pose parameters include rotation $\mathcal{R} \in SO(3)$ and translation $\mathcal{T} \in R^3$. Similar to instance-level monocular object pose detection task, the target object's CAD model is available and the target object's size $\mathcal{S} \in R^3$ is not required to be regressed. The difference is that an initial pose of the first frame is provided in instance-level object pose tracking.

\textbf{Tracking by refinement}. To the best of our knowledge, Deep 6-DOF Tracking (D6DT) \cite{garon2017deep} is the first work  that leverages deep learning for object pose tracking, which takes RGBD sequences of real objects as input. D6DT is a rendering-based method. It first uses the pose of the last frame in the sequence to render a synthetic RGBD frame for the target object. Then the deep learning network takes the rendered image and the currently observed image as input for predicting the relative 6Dof object pose between these two input frames. To improve the performance of D6DT further, Marougkas et al. \cite{marougkas2020track} integrate multiple parallel soft spatial attention modules into the network architecture to handle background clutter and occlusion. Unlike D6DT and Marougkas et al. \cite{marougkas2020track} that require taking RGBD data as input, DMB6D \cite{manhardt2018deep} takes only RGB images as input and proposes a new visual loss to drive the pose update process. DMB6D \cite{manhardt2018deep} aligns object contours in the image, hence avoiding using any explicit appearance model and depth information for pose prediction. In essence, all these three methods are implemented utilizing pose refinement, that is, they use the pose of the last frame as an initialization of the current frame, then refine the initial pose with the help of the neural networks. The final predicted pose is somehow coarse because they perform refinement simply through one single forward pass. By contrast, DeepIM \cite{li2018deepim} is proposed to refine the pose iteratively by matching the rendered image against the observed image. The relative pose between the rendered image and the observed image is predicted by a FlowNet \cite{dosovitskiy2015flownet} like network during each iteration. 

Though achieving good performance,  DeepIM needs a large amount of real-world observed images to train the network and consequently runs slow at test time. By contrast, the recently proposed SE(3)-TrackerNet \cite{wen2020se} uses only synthetic data for training yet achieves real-time performance. The superior performance of SE(3)-TrackerNet mostly comes from the rotation translation disentanglement scheme, the Lie Algebra rotation representation, and the DR data enhancement strategy \cite{tobin2017domain}.  Beyond the above methods, Busam et al. \cite{busam2020like} propose \emph{I Like to Move It}, which learns a decision process to determine an acceptable final pose iteratively. They build a decision process where an initial pose is updated in incremental discrete steps. The model sequentially moves a virtual 3D rendering towards the correct 6D pose solution. \emph{I Like to Move It}  is very similar to DeepIM and SE(3)-TrackerNet. The difference is that \emph{I Like to Move It} is designed on top of reinforcement learning, whereas the two other models are built upon deep neural networks.

In summary, existing state-of-the-art tracking by refinement methods are accurate and can achieve real-time performance, but they require the model to render CAD models into synthetic images, which is often nondifferentiable, making end-to-end training difficult to achieve. Sometimes, rendering quality and quality of the CAD models may seriously affect the tracking results.

\textbf{Tracking by optimization}. In contrast to the above tracking by refinement scheme, PoseRBPF\cite{deng2019poserbpf} achieves object tracking by using an optimized tracking scheme. Specifically, it combines Rao-Blackwellized particle filtering with a learned auto encoder network to update object pose. In PoseRBPF, a Rao-Blackwellized particle filter is used to estimate discretized distributions over rotations and translations. The translation is updated straightforward by the filter. The rotation is updated by comparing the image descriptor with precomputed entries in the codebook, generated by an auto encoder. Though PoseRBPF performs well, the continuous rotation space should be discretized in advance to obtain the codebook, which severely limits the performance of the algorithm. Moreover, the time complexity of the algorithm increases linearly as the number of discretized rotations increases. 

Majcher et al. \cite{majcher20203d} also adopt the idea of particle filtering. In their work, the target object is segmented by a U-Net in advance. Then, the 6Dof pose of the object in the current frame is estimated by matching the segmented object with the rendered image using the pose of the last frame. The final pose is obtained by optimizing the difference between object silhouettes and edges, during which the particle filter is used to estimate posterior probability distribution. PoseRBPF \cite{deng2019poserbpf} and  Majcher et al. \cite{majcher20203d} work well in a clean environment, but they are not suitable for a complicated environment such as scenarios with occlusion. Therefore, Zhong et al. \cite{zhong2020seeing} design an occlusion robust method using a so-called \emph{see through the occluder} scheme, instead of trying to detect the occluder. In Zhong et al.'s study \cite{zhong2020seeing}, a learning-based video segmentation module is integrated into an optimization-based pose estimation module to form a closed loop such that the "see through" scheme can be achieved.

Tracking by optimization methods' advantage is highly interpretable. However, they cannot achieve end-to-end training and testing because they are all partially built upon deep learning models. In addition, the optimization process is often slow. Challenges such as occlusion may also be a problem.

\subsection{Category-Level Monocular Object Pose Tracking}

Similar to instance-level object pose tracking, category-level object pose tracking also only requires the model to predict 6 degrees of freedom (6Dof) poses of a sequence of images given the initial pose. The size of the object is not required to be recovered because it can be determined by the initial pose directly. The difference is that CAD models are not allowed to be used in category-level monocular object pose tracking. Therefore, though methods such as DMB6D \cite{manhardt2018deep} claim they achieve category-level object pose tracking, they are not classified them as category-level in this study, as they use CAD models of instances in their works.

\textbf{Tracking by detection}. To the best of our knowledge, Mono3D-tracking \cite{hu2019joint} is the first work that defines the category-level 3D vehicle bounding box tracking problem and proposes to track and detect vehicles jointly in 3D from a series of monocular RGB images. Specifically, given a series of images, it first detects 2D bounding boxes of objects to generate proposals for each image. Then features of proposals are used to predict independent 3D layout (i.e., depth, orientation, dimension, and a projection of 3D center). Next, proposals of different frames are associated and refined leveraging the so-called occlusion-aware association and depth-ordering matching strategy. QD3DT \cite{hu2021monocular} adopts a similar technical routine. The difference is  QD3DT leverages quasi-dense similarity learning for object association. Moreover, 3D bounding boxes depth-ordering heuristics and motion-based 3D trajectory prediction are utilized in  QD3DT for robust instance association and occluded vehicles reidentification, respectively.

Mono3D-tracking and QD3DT achieve tracking by exploring image features, whereas 3DOT \cite{weng20203d} straightforwardly incorporates a 3D Kalman filter \cite{kalman1960new} into this task. Hungarian algorithm \cite{kuhn1955hungarian} is used in 3DOT for state estimation and data association. More specifically, 3DOT \cite{weng20203d} generates 3D bounding boxes from a series of images by an off-the-shelf 3D object detector. Then, a 3D Kalman filter is used to update the factorized parameters of these 3D boxes frame by frame. In their original paper, the 3D bounding boxes are detected from LiDAR point clouds. However, 3D bounding boxes detected from monocular RGB/RGBD images can also be used because the detecting stage and tracking stage are independent in 3DOT. Therefore, it is also classified as a monocular work. The main contribution of 3DOT is that it provides a baseline and new evaluation metrics for object pose tracking. 

Similar to 3DOT, Weng et al. \cite{weng2020joint} propose to use LSTM and graph convolutional network to update the pose, which shares the same idea with 3DOT. However, simply employing 3D Kalman filter or LSTM would cause the box drifting problem when tracking for a long range. Moreover, 3DOT and Mono3D-tracking can only track 7Dof pose, where the pitch and roll of the object are ignored. By contrast, MotionNet \cite{leeb2019motion} can recover the ignored two angles. MotionNet first detects and segments the target object at the current frame using the object mask of the last frame. Then, segmented image patches of both frames are input into a deep network to recover rotation and translation. Among the whole pipeline,  LSTM layers \cite{hochreiter1997long}  and ConvLSTM layers \cite{xingjian2015convolutional} are used to learn sequential information. Compared with other methods, MotionNet performs better in long range.

Overall, most of tracking by detection methods taking tracking as an post processing step after prediction. Therefore, they can take the advantage of most of the existing state-of-the-art detection methods. However, separating detection from tracking results in a drift of the tracking result if the sequence is very long.

\textbf{Tracking by keypoints}.
Unlike the above tracking by detection methods, CenterTrack \cite{zhou2020tracking} implements tracking by keypoints scheme that takes a pair of images as input and output object keypoints of the late frame. It adopts the framework of CenterNet \cite{duan2019centernet}. In CenterTrack, objects are localized and associated with the previous frame by predicting and associating keypoints. Specifically, it takes the image of the current frame and the last frame and the keypoints of the last frame as input, then predicts object poses of the current frame and keypoint offsets between these two frames. The object pose is tracked end-to-end through a single shot, making the algorithm fast and straightforward.However, CenterTrack can only track 7Dof object pose.  

On the contrary, Wei et al. \cite{wei2019instant} present a system for motion tracking, capable of robustly tracking planar targets and performing relative-scale 6Dof tracking without calibration, which can run in real time on mobile phones. To track planar targets, given 3D coordinates of 4 objects corners in the previous frame and the corresponding 2D coordinates in the current frame, they solve the rigid body transformation (3D rotation and translation) using a Levenberg Marquardt optimization algorithm. Later, Ahmadyan et al. \cite{ahmadyan2020instant} improve the work of Wei et al. \cite{wei2019instant} by tracking the projected 2D coordinates of the 3D bounding boxes' 9 keypoints (8 corners and 1 center). After all keypoints are tracked, the up-to-scale 6Dof pose can be recovered by solving an EPnP problem. 

Though lightweight and simple, both methods \cite{wei2019instant,ahmadyan2020instant} have a common defect, they can only estimate the up-to-scale 6Dof pose, which is viable for AR/VR application yet is not sufficient for robotic grasping or autonomous driving. Therefore, 6-PACK \cite{wang20206} is proposed to take monocular RGBD images as input and estimate object pose by accumulating relative pose changes over time. More specifically, an anchor-based keypoint-generation scheme is first employed to generate keypoints adaptively from the previous frame and the current frame. Then, two sets of ordered keypoints, as well as previously computed instance poses, are used to obtain the current estimated 6Dof pose. The main limitation of 6-PACK is that it can only work when depth information is available and cannot generalize to RGB-only scenarios.

In conclusion, tracking by keypoints is advanced because they are lightweight and simple. The detection and tracking process are often unified. However, they are limited by lacking freedom, up-to-scale solution and dependent on depth.

Till now, we have introduced the recent state-of-the-art monocular object pose tracking methods. Providing a unified quantitative comparison is difficult because they focus on different scenarios and take different data formats as input.

\section{Analysis for Possible Future Works and Challenges}
\label{secfuturework}
This work has introduced the latest developments in monocular object pose estimation from the perspectives of instance-level detection, category-level detection and object pose tracking. Next,  to help the community develop this research direction better, several possible future research directions and challenges are analyzed and introduced in this section.

\textbf{For instance-level object pose detection using RGB images only}, the following are observed:
\begin{itemize}
	\item First, though existing algorithms have performed sufficiently well in simple indoor scenarios, they have difficulty handling  situations where occlusion, truncation and cluttered backgrounds exist.  However, interference such as occlusion is inevitable in practical applications. Therefore, studying how to deal with complex interference such as occlusion is a good future research direction. A possible reason for their poor performance is that most current methods rely on keypoints that should be uniformly sampled from all viewpoints of the objects, for example, keypoints sampled by furthest sampling. This reason may cause the trained model be over-fitted to uniformly sampled keypoints, which cannot be predicted when occlusion or truncation exists.

	\item Second, existing RBG-only methods are very susceptible to factors such as lighting changes and shooting angles. These factors can cause image blur, reflection, blind spot, and cutoff, which may obscure features extracted from images,  especially when these features are used for detecting keypoints. This limitation may not be a great deal for environment-controlled indoor scenarios (e.g., indoor  factories). However, in outdoor applications, such as AR on mobile phones, this would become the largest obstacle for its wide application because the lighting condition is uncontrollable and unpredictable. Therefore, designing  algorithms that are robust to the above factors should also be an important research topic in the future. A straightforward solution is to design more robust backbone networks to learn deep features, but achieving it is difficult and deep networks' interpretability is poor. Another possible solution is to build larger datasets to make the network to learn such kinds of scenarios, though it would be costly.  Moreover, adopting transfer learning and domain adaptation in 6D object pose detection can solve part of this problem.

	\item Third, existing research has shown that establishing 2D-3D correspondences for object pose estimation performs better than directly predicting pose parameters. Hence, mainstream works are long committed to research on how to establish correspondences better. However, this kind of method cannot be trained in an end-to-end manner, and building and solving correspondence relations is time-consuming. Therefore, future works need to consider designing differentiable 2D-3D correspondences solving algorithms applied in the neural networks or exploring the possibility of improving the performance of correspondence-free methods. For example,  several deep network architectures can be designed to simulate the process of the PnP algorithm.  The overall pose detection process would be differentiable and end-to-end training can be achieved.
	
	\item Lastly, the output of existing methods has a very violent jitter because the pose is predicted from each frame independently. The problem can be solved by taking multiple-frames as input. Another possible solution is to add post-processing processes to prevent jitter, for example, median or mean filter. However,  an appropriate threshold for filtering needs to be selected, otherwise it will produce a visual floating effect.
\end{itemize}

\textbf{For instance-level object pose detection using (RGB)D data}, the following are observed:
\begin{itemize}
	\item  Though existing methods always perform much  better than RGB-only methods, they often consume much more computational resources because additional depth information needs to be learned, and several methods require additional refinement steps such as ICP for performance gain, which further increases running time. Therefore, designing a more lightweight network architecture to reduce time and space complexity may be a valuable future research topic. A possible reason for the slow running time is that the point cloud processing network requires searching for the K nearest neighbors to learn local features. Utilizing sparse voxel convolutional networks may be a feasible solution.
	
	\item  Most existing low-power hardware such as mobile phones can only capture sparse point clouds in use.  While existing (RGB)D-based methods are all evaluated on datasets with dense point clouds generated from depth maps, their performances on sparse point clouds are under explored, and this has caused a bias between evaluating performance and practical use. Therefore, studying whether current algorithms are suitable for taking sparse point clouds as input or not is necessary. If not, tailored new algorithms should be proposed. 
	
	\item Labeling 6Dof pose of objects is challenging. Therefore, another important challenge is how to obtain precise ground-truth poses. Owing to the existing advanced computer graphics technologies, synthetic data with ground truth are very easy to obtain, and can be used for object pose detection training. Nevertheless, models trained on synthetic datasets usually perform poorly on real-world images. Therefore, this brings up a new possible future research question: how to improve the generalization ability of models trained on synthetic datasets. Existing self-supervised learning methods have provided some promising pre-research, but more efforts is needed. This research direction is also suitable for both RGB-based instance-level methods and category-level methods.
\end{itemize}

In contrast to instance-level monocular object pose detection, for category-level monocular object pose detection, category-level monocular 3D object detection and category-level monocular 6D pose detection are relatively niche.

\textbf{For category-level 3D object detection, the following are observed}

\begin{itemize}
	\item As its main application provides environmental information for autonomous driving, locating objects is more important than predicting the size and orientation of objects. However, locating objects in the 3D space using a single RGB image is ill-posed. Therefore, equipping the model with the ability to predict depth is essential. Given that images utilized by this task usually contain multiple objects and contain a wide range of feature-rich backgrounds,  Using them to infer depth information may be a feasible solution, that is, utilizing an instance-aware relationship to improve the model's depth perception ability should be researched, especially for utilizing non-local features hidden in images. Incorporating Vision-Transformers \cite{liu2021swin,wang2021pyramid} into the network architecture may be a good idea.
	
	\item Utilizing Pseudo-LiDAR is a feasible research direction. However, current pseudo-LIDAR-based solutions usually use off-the-shelf depth prediction models to predict depth in advance. It has caused gaps between 3D detection and depth prediction that is, current depth estimation models suffer from the sub-optimal problem, and utilizing Pseudo-LiDAR point clouds generated by them for 3D detection would further aggravate the problem. Therefore, in the future research of pseudo-LiDAR, combining depth estimation and 3D detection into one unified network or into the same training process to obtain mutual performance gain and avoid detection errors accumulated by different sub-optimal problems may be valuable.  
	
	\item Existing datasets such as KITTI3D always contain  point clouds captured by LiDAR and images captured by a monocular camera. Though these point cloud data are not allowed to be used in monocular detection tasks at inference time, studying how to utilize them better for training a monocular 3D object detector is meaningful. For example,  point cloud can be utilized to learn convolutional weights during training while discarding them during inference. Alternatively off-the-shelf point cloud 3D detectors as teacher networks to train monocular 3D detectors can be used, as has been done in knowledge distillation \cite{chen2017learning,kim2016sequencee}.

\end{itemize}

\textbf{For category level monocular 6D pose detection},  shortcomings of existing methods are clear:

\begin{itemize}
	\item Most of the methods need to use an off-the-shelf 2D object detection model to locate the target object in advance. Then the target image patch is cropped out and resized before performing pose prediction. Such a two-stage scheme may cause the accumulation of locating errors. Therefore, whether generating object proposals and completing pose estimation in a unified network or through a totally proposal-free way is possible is a question. The answer is clearly yes, referring to the successful experience of anchor-free 2D object detection models \cite{duan2019centernet,zhu2019feature}. However, to date, no researchers have worked in this direction. 
	
	\item Existing methods always use large backbones, such as ResNet-101, as feature extractors, which is instrumental in improving effectiveness but is conducive to improving efficiency.  Coupled with the time-consuming 2D object detection process, guaranteeing whether the model can achieve real-time performance or not is difficult. Therefore, how to design lightweight real-time models is worth studying in the future. As the first step,  whether existing light-weight backbones like MobileNet \cite{sandler2018mobilenetv2} and ShuffleNet \cite{zhang2018shufflenet} are suitable for category-level monocular 6D pose detection should be researched. Besides, how to apply lightweight model methods like knowledge distillation, quantification, and pruning to this task should also be investigated.
	
	\item Most of the existing algorithms are highly dependent on utilizing depth information. We note that the RGB data are equally important. If both data are available,  more attention should be paid to how to fuse features learned from RGB data and depth data. Transformer, graph convolutional networks, and gated memory networks are all possible research directions. Moreover, taking only RGB images as input in daily applications, such as  augmented reality-related apps on mobile phones is more practical. The performance of the RGB-only methods shows great potential in future studies.Estimating the depth from RGB input or studying monocular scene reconstructing  may be a possible solution. In addition, directly predicting pose related parameters by adopting 2D object detection  pipelines  is possible, in view of the successful experience of monocular 3D object pose detection.
	
\end{itemize}

\textbf{For monocular object pose tracking}, the following are noticed:

\begin{itemize}
	\item If the CAD model is available,  solving this problem is not difficult in a controlled scenario. In uncontrollable scenes (such as autonomous driving scenes and outdoor lighting scenes), all the problems in the instance-level object pose detection task will be faced. 
	
	\item Existing object pose tracking algorithms usually only take two frames of images (the current frame and one previous frame) as input to predict the object poses of the current frame, and this could cause three main problems: first, the sequential information is not fully utilized. Second, tracking errors accumulate over time and cannot be eliminated. Third, the box drift problem may occur. To solve these problems, monocular object pose tracking introduces a feasible future research direction, that is, using recurrent neural networks such as LSTM for associating multi-frame information. It can not only improve the utilization of features but also ensure the stability of the tracking results. 
	
	\item Many existing methods need to render CAD models, which is time-consuming, as most existing renderers are either non-differentiable or cost-in-efficient. Therefore, designing efficient and differentiable rendering algorithms is essential in future works.Researching how to utilize Nerf \cite{yen2020inerf} for object pose tracking may be a meaningful research direction.
	
	\item In addition, when the CAD model is not available, most of the existing works only track 7Dof 3D bounding boxes. To the best of our knowledge, only one work \cite{wang20206} can achieve full 9Dof category-level pose tracking. As mentioned before, 7Dof pose is sufficient for location-aware scenarios such as autonomous driving but not for rotation and size-aware scenarios such as augmented reality. Therefore, future research should pay more attention to tracking the full 9Dof bounding boxes.

\end{itemize}

\textbf{For research on datasets}, attention can be paid on following aspects:
\begin{itemize}
	\item Considering the difficulty of obtaining real-world ground truth, future studies should focus on designing open-source full 9Dof pose annotation tools, especially tools available on category-level video annotation.
	
	\item Most of the existing datasets place codermarkers in the scene to help obtain ground truth poses. However, these codermarkers may provide extra characteristics that would never exist in practical application scenarios. Hence, the bias would occur when evaluating and deploying models trained on these datasets. Therefore, real-world datasets that consist of images without codemarkers should be captured and released.  
	
	\item For category-level monocular object detection, intra-class variation  between instances of the same category in the current dataset NOCS is still limited, that is, instances are not rich enough, which may make models trained on them hard to generalize to new instance that are quite different to them in appearance. Therefore, the number of distinct instances within the same category should increase for both training and testing to increase intra-class variation for current datasets.  
	
	\item Although several works have been proposed for object pose tracking, a unified benchmark dataset to evaluate and fairly compare them is still lacking. Therefore, collecting and releasing a large-scale dataset for benchmarking object pose tracking methods is necessary. 
	
\end{itemize}

\section{Conclusion}
\label{secconclusion}
This study presents a contemporary survey of the state-of-the-art deep learning-based methods for monocular object pose detection and tracking, including instance-level monocular object pose detection, category-level monocular object pose detection, and monocular object pose tracking. Metrics and datasets are introduced in detail. Moreover, a comprehensive taxonomy and performance comparison of these methods are presented qualitatively and quantitatively. Furthermore, the potential research directions and challenges are analyzed in the end.

\begin{acks}
This work was supported by the National Key Research and Development Program of China under Grant No.2020YFB2104100 and the National Natural Science Foundation of China (NSFC) under Grant No. 62172421 and 62072459. And we would like to thank Prof. Chunhua Shen from the University of Adelaide for his help of improving this paper.
\end{acks}

\bibliographystyle{ACM-Reference-Format}
\bibliography{mybibfile}


\end{document}